%% file: PAN.tex
\documentclass[acmlarge]{acmart}

\AtBeginDocument{%
  \providecommand\BibTeX{{%
    \normalfont B\kern-0.5em{\scshape i\kern-0.25em b}\kern-0.8em\TeX}}}

\setcopyright{acmcopyright}
\copyrightyear{2019}
\acmYear{2019}
\acmPrice{15.00}
\acmDOI{10.1145/3369816}

\acmJournal{IMWUT}
\acmVolume{3}
\acmNumber{4}
\acmArticle{144}
\acmMonth{12}

\usepackage{hyperref}
\usepackage{url}
\usepackage{graphicx}
\usepackage{enumerate}

\usepackage[caption=false,font=normalsize]{subfig}
\usepackage{xspace}
\usepackage[ruled]{algorithm2e}
\usepackage{multirow, makecell}
\usepackage{enumitem}

\newcommand{\ie}{\emph{i.e.},\xspace}
\newcommand{\eg}{\emph{e.g.}\xspace}

\newcommand{\etal}{\emph{et al.}\xspace}

\newcommand\figref[1]{Figure~\ref{#1}}

\newcommand\tabref[1]{Table~\ref{#1}}

\newcommand\equref[1]{Eq.(\ref{#1})}

\newcommand{\systemname}{{\sf PAN}\xspace}
\newcommand{\systemnameposs}{{\sf PAN's}\xspace}

\newcommand\rev[1]{\textcolor{black}{#1}}

\newcommand\lsc[1]{\textcolor{black}{#1}}

\begin{document}

\title{Privacy Adversarial Network: Representation Learning for Mobile Data Privacy}
% \title{Better Accuracy with Quantified Privacy: Representations Learned via Privacy Adversarial Network}

\author{Sicong Liu}
\affiliation{%
  \institution{Xidian University}
  \department{School of Computer Science and Technology}
  \city{Xi'an}
  \country{China}
}
\author{Junzhao Du}
\authornote{Corresponding Author: Junzhao Du}
\affiliation{%
  \institution{Xidian University}
  \department{School of Computer Science and Technology}
  \city{Xi'an}
  \country{China}
}
\author{Anshumali Shrivastava}
\affiliation{%
  \institution{Rice University}
  \department{Department of Computer Science}
  \city{Houston}
  \state{TX}
  \country{USA}
}
\author{Lin Zhong}
\affiliation{%
  \institution{Rice University}
  \department{Department of Electrical $\&$ Computer Engineering}
  \city{Houston}
  \state{TX}
  \country{USA}
}

% \author{Anonymous Author(s)}
% \affiliation{%

% \affiliation{%

% \affiliation{%

% \affiliation{%
% %  \country{India}}

% \affiliation{%

% \affiliation{%




\begin{abstract}
The remarkable success of machine learning has fostered a growing number of cloud-based intelligent services for mobile users. Such a service requires a user to send data, \eg image, voice and video, to the provider, which presents a serious challenge to user privacy. 
To address this, prior works either obfuscate the data, \eg add noise and remove identity information, or send representations extracted from the data, \eg anonymized features. They struggle to balance between the service utility and data privacy because obfuscated data reduces  utility and extracted representation may still reveal sensitive information.

This work departs from prior works in methodology: we leverage adversarial learning to better balance between privacy and  utility.
% \rev{on the generated representations}. 
%
We design 
%\lin{revisit ``design''} 
a \textit{representation encoder} that generates the feature representations to optimize against the privacy disclosure risk of sensitive information (a measure of privacy) by the \textit{privacy adversaries}, and concurrently optimize with the task inference accuracy (a measure of utility) by the \textit{utility discriminator}. 
The result is the privacy adversarial network (\systemname), a novel deep model with the new training algorithm, that can automatically learn representations from the raw data. 
%\lin{Lin revisit these two senetnces}
%
And the trained encoder can be deployed on the user side to generate representations that satisfy the task-defined utility requirements and the user-specified/agnostic privacy budgets.
% generate representations, that are both private and useful.
% for the task inference services.

Intuitively, \systemname adversarially forces the extracted representations to only convey information required by the target task. Surprisingly, this constitutes an implicit regularization that actually improves task accuracy. 
As a result, \systemname achieves better utility and better privacy at the same time! We report extensive experiments on six popular datasets, and demonstrate the superiority of \systemname compared with alternative methods reported in prior work.
\end{abstract}


% \keywords{XXX}

% \keywords{mobile data privacy, adversarial network, gaze detection, text tagging}

\maketitle
\renewcommand{\shortauthors}{Liu et al.}

\input{body/intro}
\input{body/formulation}

\input{body/design}
\input{body/experiment}
\input{body/interpretation}

\input{body/related}
\input{body/conclusion}
\begin{acks}
This work is supported in part by National Key R$\&$D Program of China $\#2018YFB1003605$, Natural Science Foundation of China (NSFC) $\#61472312$, Shaanxi Fund $2018JM6125$, Open Fund of State Key Laboratory of Computer Architecture, The Youth Innovation Team of Shaanxi Universities, and Natural Science Foundation (NSF) Grant $\#1611295$. 
The idea behind \systemname was conceived during Sicong Liu's yearlong visit to Rice University with support from China Scholarship Council to which the authors are grateful.
The authors also thank the anonymous reviewers for their constructive feedback that has made the work stronger. 
% \lscrev{Sicong Liu is grateful to China Scholarship Council, Professor Lin Zhong, and Professor Anshumali Shrivastava for their support and guidance to start this work at Rice University.}
% This work is partially supported by the National Natural Science Foundation of China (NSFC) under Grant No. 61502374, 61472312, and 61272456; the Fundamental Research Funds for the Central Universities under project No. BDY041409 and JB151002 (Xidian University); and the CETC shining Star Innovation. \footnote{Article Number: 17}
\end{acks}

\bibliography{acmart}
\bibliographystyle{ACM-Reference-Format}

\end{document}

%% file: body/intro.tex
\section{introduction}
\label{sec:introduction}
% \rev{Compared with other published papers in ICLR, it is better to delete half page of Introduction.}
% \rev{sc's hint1: deliver raw data to service provider is not safe.}
% \textbf{Motivation.}
Machine learning has benefited numerous mobile services, such as speech-based assistant (\eg Siri), reading log enabled book recommendation (\eg Youboox).
%\lin{FaceID is not a pertinent example because the computation is done locally without sending data to the cloud. Please update}
% \lin{we are not addressing the privacy of TRAINING DATA. We are addressing the privacy when users use the trained models (by submitting raw data)}
Many such services submit user data, \eg sound, image, and human activity records, to the \textit{service provider}, posing well-known privacy risks~\cite{bib:abadi2016:abadi, bib:dwork2017:ARSA, bhatia2016privacy}.
%
% For example, with the collected sensor data, malicious party may conduct the underlying correlation detection, re-identification and other malicious mining~\cite{bib:dwork2017:ARSA, bhatia2016privacy}.
%
% \TODO{Double check the following paragraphs after finishing the other sections:}
% Different from pinning hopes on service providers to anonymize data for privacy-preserving, 
Our goal is to avoid disclosing raw data to service providers by \lsc{creating a device-local intermediate component that encodes the raw data and only sends the encoded data to the service provider.}
And the encoded data must be both \textit{useful} and \textit{private}.
For inference-based services, \textit{utility} can be quantified by the inference accuracy, achieved by the service provider using a discriminative model.
And \textit{Privacy} can be quantified by the 
disclosure risk of private information.

% can be quantified by the inference accuracy, achieved by the service provider using a discriminative model.
% which is the inverse of \emph{discriminative error}.

%
% First, the collective raw images, audio, and personal activity data often accompany abundant sensitive information of mobile users and others in their social circle, such as faces, speeches, context image, ambient noise, activity timetable, etc.
% %
% Second, mobile users who authorize data collection for an inference service can neither recall it, nor control how, where and by whom it will be maliciously mined.
% %
% Third, although the service provider state that the collective data from massive mobile users is well protected by anonymous techniques~\cite{bib:sweeney2002:UFKBS, bib:islam2011:KBS}, or release them in a statistic form using differential privacy techniques~\cite{bib:dwork2014:FTTCS, bib:shokri2015:sigsag}, but the original data is still kept by service providers and more services derived from these data are also monopolized by the service provider.
% %
% Moreover, service provider is not trustworthy once mobile users' raw data flow into their packets,
% %
% for instance, the big data leak events at Yahoo in 2013, at eBay in 2014, at Anthem in 2016 and at Facebook in 2018~\cite{url:dataleak}.

% \textbf{Prior work and challenges.} 
Existing solutions addressing the privacy concern struggle to balance between above two seemingly conflicting objectives: privacy vs. utility.
An obvious and widely practiced solution is to transform the raw data into task-specific features and upload features only, like Google Now~\cite{url:googlenow} and Google Cloud~\cite{url:googlecloud};
% Google Cloud Machine Learning Engine also provides API to preprocess the raw data into engineering features before uploading~\cite{url:googlecloud}.
This not only reduces the data utility but also is vulnerable to 
reverse models that reconstruct the raw data from extracted features~\cite{bib:mahendran2015:CVPR}.
% \rev{anshu: This is because the good features learned by either shallow feature engineering techniques or deep learning algorithms remove this} \rev{try to extract the essence of raw data, which contributes to inference accuracy in subsequent classification but incurs easy data reconstruction by reverse techniques~\cite{bib:rifai2011:ICML}}.
% \lin{``which....'' does not make any sense to me}
The authors of \cite{bib:ossia17:arxiv} additionally apply dimensionality reduction, Siamese fine-tuning, and noise injection to the features before sending them to the service provider. This unfortunately result in further loss in utility.

% \textbf{Our design.}
Unlike previous work, we employ deep models and adversarial training to automatically learn features for a sweet tradeoff between \textit{privacy} and \textit{utility}.
Our key idea is to judiciously combine the discriminative learning, for minimizing the task-specific discriminative error as well as maximizing the user-specified privacy discriminative error, and the generative learning, for maximizing the agnostic privacy reconstruction error.
Specifically, we present the Privacy Adversarial Network (\systemname), an end-to-end deep model, and its training algorithm.
\systemname controls three types of descent gradients, \ie utility discriminative error, privacy discriminative error, and privacy reconstruction error, in back propagation to guide the training of a feature extractor.

As shown in Fig.~\ref{fig_ran_design}, a \systemname consists of four parts: a feature extractor (Encoder $E(\cdot)$), a utility discriminator (UD), an adversarial privacy reconstructor (PR), and an adversarial privacy discriminator (PD).
The output of the Encoder (E) feeds to the input of the utility discriminator (UD), privacy reconstructor (PR), and privacy discriminator (PD).
We envision the Encoder (E) runs in mobile devices to extract features from raw data. The utility discriminator (UD) represents the inference service to ensure the utility of extracted features. 
\systemname emulates two types of adversarials to ensure the privacy:
the privacy discriminator (PD) emulates a malicious party that seeks to extract private information, \eg user location;
the privacy reconstructor (PR) emulates one that seeks to reconstruct raw data from the features.
%
% There is no theoretic guarantee on end-to-end training the colloborated discriminative model and generative model for straightforward feature extracting, or feature disentangling.
%
We present a novel algorithm to explicitly train \systemname via an adversarial process that alternates between \ie training the Encoder with the utility discriminator (UD) to improve the utility and confronting the Encoder with the adversaries of privacy discriminator (PD) and privacy reconstructor (PR) to enhance the privacy.
All four parts iteratively evolve with others during the training phase.
% from each other.
%
Understood from the perspective of manifold, the separate flows of gradients 
from utility discriminator (UD), privacy discriminator (PD), and privacy reconstructor (PR) through the Encoder in back-propagation can iteratively produces the feature manifold that is both useful and private.

Using digit recognition (MNIST~\cite{data:mnist1998:LeCun}), image classification (CIFAR-10\cite{data:cifar} and ImageNet~\cite{data:imagenet}), sound sensing (Ubisound~\cite{bib:sicong2017:IMWUT}),  human activity recognition (Har~\cite{data:Har}), and driver behavior prediction (StateFarm~\cite{data:statefarm}), we show \systemname is effective in training the Encoder to generate deep features that provide better privacy-utility tradeoff than other privacy preserving methods. 
Surprisingly, we observe that the adversarially learned features to remove redundant information, for privacy, even surpass the recognition accuracy of discriminatively learned features. 
%\lin{this sentence does not make sense. What is the original model? How could you compare ``features'' with ``model''? Did you actually mean that the adversarially trained Encoder+UD is better than a discriminatively trained deep model? } \lscrev{Yes, it is want I mean.}
That is, removing task-irrelevant information for privacy actually improves generalization and as a result, utility.

In the rest of the paper, we formulate the problem of utility-privacy tradefoff in $\S$ \ref{sec:problem} and present the \systemnameposs design and its training algorithm in $\S$ \ref{sec:pan}.
We report an evaluation of \systemname in $\S$ \ref{sec:experiment}. 
We attempt a theoretic interpretation of \systemname in $\S$~\ref{sec:discuss}, review the related work in $\S$ \ref{sec:related}, and conclude in $\S$ \ref{sec:conclude}. 

%% file: body/formulation.tex
\section{Problem Definition of Mobile Data Privacy Preserving}
\label{sec:problem}

\begin{figure}[t]
\centering
\includegraphics[width=.8\textwidth]{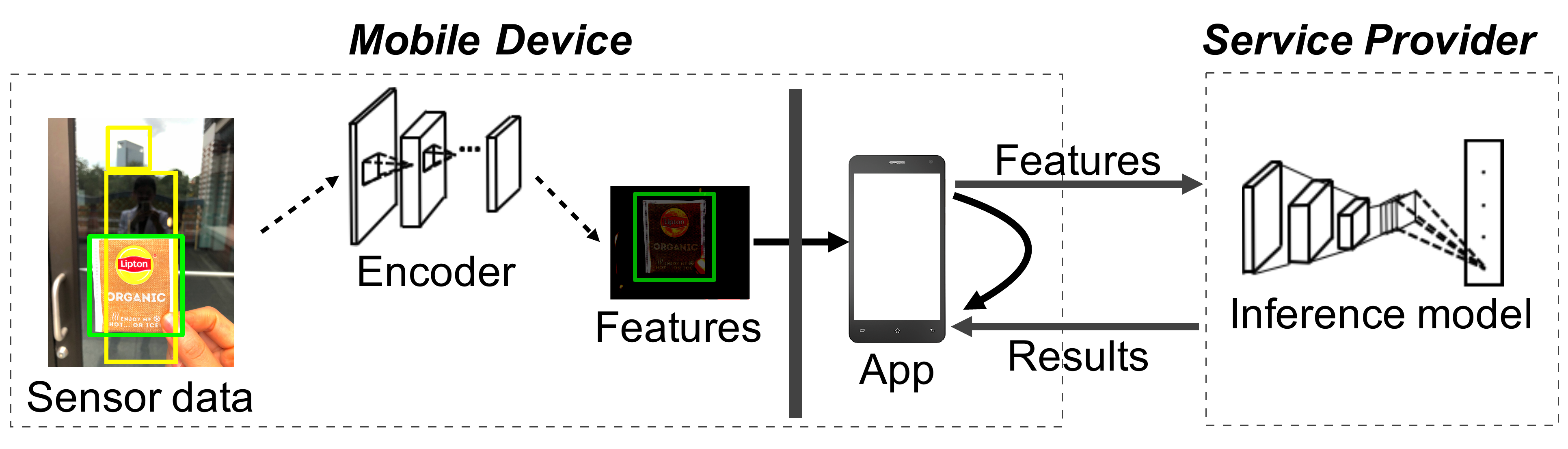}
%\vspace{-3mm}
\caption{Privacy preservation in cloud-based mobile services. Mobile users leverage a learned Encoder to locally generate deep features from the raw data (\ie "tea bag" picture) and give them to the App. The App may send the features to its cloud-based backend.}
%\vspace{-4mm}
\label{fig_privacy_program}
\end{figure}

This section mathematically formulates the problem of utility-privacy tradeoff for mobile data.
Many appealing cloud-based services exist today that require data from mobile users.
For example, as shown in Fig \ref{fig_privacy_program}, a user takes a picture of a product and sends it to a cloud-based service to find out how to purchase it, a service Amazon actually provides.
The picture, on the other hand, can accidentally contain sensitive information, e.g., face and other identifying objects in the background.
Therefore, the user faces a touch challenge: how to obtain the service without trusting the service provider with the sensitive information?

Toward addressing this challenge, our key insight is that most services actually do not need the raw data. The user can encode raw data $I$ into representation $E(I)$ through an Encoder $E(\cdot)$ on the mobile device and only sends $E(I)$ to the service provider.
The representation $E(I)$ ideally should have the following two properties:
 \begin{itemize}
 \item 
\textbf{Utility:} it must contain enough task-relevant information to be useful for the intended service, e.g., high accuracy for object recognition;
 \item
\textbf{Privacy:} it must have little task-irrelevant information, especially that is considered sensitive by the user.
 \end{itemize}
%
% All of the Encoder, Decoder and Classifier are trained by service provider, 

%\subsection{Utility and Privacy Metrics}
In this work, we focus on classification-based services. Therefore, the \textit{utility} of $E(I)$ is measured by the task inference error $C_u$ (\eg cross entropy) in the service provider.
And we quantify the \textit{privacy} of $E(I)$ by the privacy leak risk $C_p$ of raw data in all possible attacking models $X$.
%an intuitive metric, \ie the reconstruction error in a reversed deep model, $X$, employed by a malicious party.
%The reconstruction error measures the risk of original data disclosure.
%
Since the Encoder is distributed to mobile users, we assume it is available to both service providers and potential attackers. That is, both the service provider and the malicious party can train their models using raw data $I$ and their corresponding Encoder output $E(I)$.
As such we can restate the desirable properties for the Encoder output $E(I)$ within dataset $\mathbf{T}$ as below: 
\begin{equation}
\label{equ_objective}
\begin{split}
\text{\textbf{Utility}:} \quad & \underset{E}{Min} \  C_{u}(E(I_i)), i \in \mathbf{T} \\
\text{\textbf{Privacy}:} \quad & \underset{E}{Min} \ \underset{X}{Max} \  C_{p}(E(I_i)),i \in \mathbf{T} \\
% & Min \ Obj3= \lambda \mathcal{H} (y,y')-(1-\lambda) |I-I'|^2 \\
\end{split}
\end{equation}
%where, $U(E(I_i))$ denotes the inference loss with the testing data $i \in \mathbf{T}$. And $P(E(I_i))$ is the privacy leak risk of raw data given the Encoder output $E(I)$.
%$Y_i'$ and $Y_i$ is the inference class and the true label, respectively.
%
%$|I_i-I_i'|^2$ is the  Euclidean distance, \ie reconstruction error, between raw data $I_i$ and the mimic data $I_i'$ reconstructed by a malicious party with the Encoder output.
%

The first objective (\textbf{Utility}) is well-understood for discriminative learning, and achievable via a standard optimization process on the Encoder (E) and the corresponding specialist discriminative model, \ie minimizing the cross entropy between the predicted task label and the ground truth in a supervised manner~\cite{bib:kruse2013:CI}. 

The second objective (\textbf{Privacy}) has two parts. The inner part, $\underset{X}{Max}\ C_p(E(I))$, is 
opposite to the the outer part $\underset{E}{Min}\ C_p(E(I))$.
Therefore, the Encoder ($E$) employed by the mobile user and the specialist attacker ($X$) used by the malicious party is adversarial to each other in their optimization objectives.
Given the information loss in $E(I)$ for privacy, utility loss appears to be certain in theory.
% of the app and vice-a-versa.
%
One would only hope to find a good, ideally Pareto-optimal, tradeoff between privacy and utility in devising $E(\cdot)$.
However, as we will show later, $E(\cdot)$ discovered via \systemname actually improves privacy and utility at the same time, a result that can be explained by the practical limits of deep learning in \S\ref{sec:discuss}.  

Important to the quantification of privacy, one must enumerate the privacy leak risks $C_p$ by all possible attackers $X$ in theory. Moreover, the measurement of the privacy leak risk $C_p$ is an open problem in itself~\cite{bib:mendes2017:access}.
Therefore, we approximate privacy with two specific attackers, each with its own measurement of privacy, elaborated below. 
\begin{enumerate}

\item \textit{Specified privacy quantification} in which the user specifies what inference tasks should be forbidden and privacy can be quantified by the accuracy of these tasks.
For example, users may want to prevent a malicious party from inferring their identity. In this case, the privacy can be measured by the inaccuracy of identify inference.  
In this case, the privacy leak risk $C_{p1}$ can be defined as the inference accuracy by a discrimination model employed by the attacker.
%with the exactly identified privacy class.
%
\item \textit{Intuitive privacy quantification} in which the privacy leakage risk $C_{p2}$ is agnostic of the inference tasks under taken by the attacker.
In this work, we quantify this agnostic privacy by the difference between the raw data, $I$, and $I'$, data reconstructed by a malicious party from the Encoder output $E(I)$.
We choose this reconstruction error as the agnostic measure for two reasons.
First, the raw data in theory contains all information and difference between $I$ and $I'$ is computationally straightforward and intuitive.
Second, prior works have already shown that it is possible to reconstruct the raw data from feature representations optimized for accuracy~\cite{bib:mahendran2015:CVPR, bib:radford2015:arxiv, bib:zhong2016:FDS}.
\end{enumerate}

%% file: body/design.tex
\section{Design of PAN}
\label{sec:pan}
To find a good, hopefully Pareto-optimal tradeoff between utility and privacy, we design \systemname to learn an Encoder $E(\cdot)$ via a careful combination of discriminative, generative, and adversarial training. As we will show in \S\ref{sec:experiment}, to our surprise, the resulting Encoder actually improves utility and privacy at the same time.
%To tackle above challenges, we present \systemname to train a feature extractor, \ie Encoder, with good trade-offs between privacy and utility in \equref{equ_objective}.
% To learn an Encoder (E) for the utility-privacy objective (eq. \eqref{equ_objective}), this section introduce the design of \systemnameposs.

\subsection{Architecture of PAN}
\label{subsec:arc_pan}

\begin{figure}[t]
\centering
\includegraphics[width=.95\textwidth]{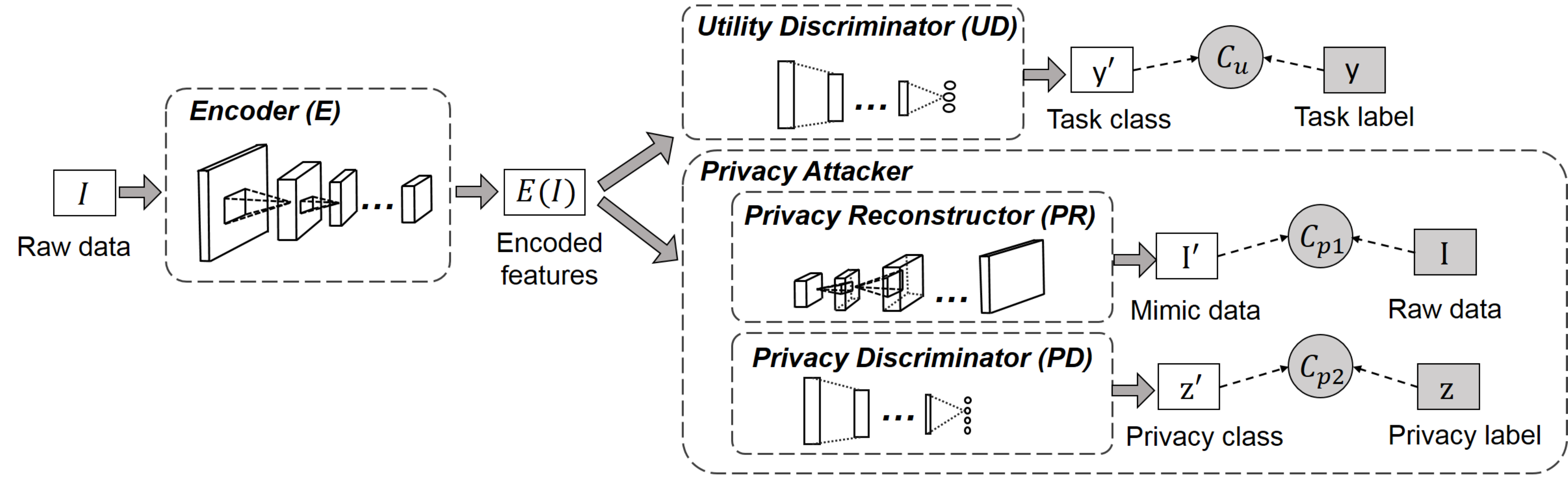}
%\vspace{-4mm}
\caption{Architecture of the privacy adversarial network (PAN). The error $C_u$, $C_{p1}$, and $C_{p2}$, which respectively refer to the utility quantification, the specified privacy quantification and the intuitive privacy quantification, build the weight updating criterion during back-propagation.}
% The raw data $I$ inputs an Encoder, and then the generated features $E(I)$ feed into a separate Classifier and Decoder for reconstructed data $I'$ and inference results $Y'$, respectively. The descent gradients from three optimization functions (\ie $O_d$, $O_g$, $O_a$) build the weight updating criterion during back-propagation.}
\label{fig_ran_design}
%\vspace{-4mm}
\end{figure}

As shown in Fig \ref{fig_ran_design}, \systemname employs two additional neural network modules, utility discriminator (UD) and privacy attacker, to quantify utility and privacy, respectively, in training the Encoder $E(\cdot)$.
The utility discriminator simulates the intended classification service; when \systemname is trained by the service provider, the utility discriminator can be the same discriminative model used by the service. 
The privacy attacker, \ie the intuitive privacy reconstructor (PR) and the specified privacy discriminator (PD), simulates a malicious attacker that attempts to obtain sensitive information from the encoded features $E(I)$.
These modules are end-to-end trained to learn the Encoder $E(\cdot)$ for users to extract deep features $E(I)$ from raw data $I$.
% on mobile and embedded platforms.
% 
% We note that when property compressed, the Encoder can fit into resource-constrained mobile devices~\cite{bib:liu2018:mobisys}.
% 
The training is an iterative process that we will elaborate in \S \ref{subsec:algorithm}. Below we first introduce \systemnameposs neural network architecture, along with some empirically gained design insights. 

\begin{itemize}[leftmargin=*]
\item The \textbf{Encoder $E(\cdot)$} consists of an input layer, multiple convolutional layers, pooling layers, and batch-normalization layers. 
The convolution layer applies a convolution operation to output activation map with a set of trainable filters.
We note that the clever usage of pooling layers and batch-normalization layers contribute to the deep feature's utility and privacy.
The batch-normalization layer normalizes the outputted activation map of a previous layer by subtracting the batch mean and dividing by the batch standard deviation~\cite{bib:ioffe2015:arxiv}.
It helps the features' utility because it normalizes the activation to avoid being too high or too low thus has a regularization effect~\cite{bib:ioffe2015:arxiv}.
It contributes to features' privacy as well since it makes it harder for an attacker to recover sensitive information from normalized features.
And then, the pooling layer adopts a maximum or average value from a sub-region of the previous layer to form more compact features, which reduces the computational error and avoids over-fitting~\cite{bib:giusti2013:ICIP}. 
It helps privacy because none of un-pooling techniques can recover fine details from the resulting features through shifting small parts to precisely arranging them into a larger meaningful structure~\cite{bib:milletari2016:3DV}.

\item The \textbf{Utility Discriminator (UD)} builds a multi-layer perceptron (MLP) to process deep features $E(I)$ and output the task classification results $y'$ with several full-connected layers~\cite{bib:kruse2013:CI}.
% of non-linearly-activating nodes.
We also note that a service provider can explore any classification architectures for its utility discriminator model, given the Encoder or its binary version. 
We choose the MLP architecture because some of the most successful CNN architectures, \eg VGG and AlexNet, can be viewed as the Encoder plus an MLP.
The standard cross entropy between the utility discriminator's prediction output $y'$ and the task ground truth $y$ measures the utility error $C_u$.

\item The \textbf{Privacy Attacker} employs the two privacy attacking models presented at the end of \S\ref{sec:problem}. Specifically, the privacy discriminator (PD) evaluates the recognition accuracy of private class from encoded features $E(I)$. 
And the privacy reconstructor (PR) quantifies the intuitive reconstruction error between mimic data $I'$ and raw data $I$.
\begin{itemize}[itemindent=2em]

\item \textbf{Specified Privacy Discriminator (PD)} employs a similar MLP classifier as the utility discriminator (UD) to predict the user-specified privacy class $z'$, \eg personal identity, from features $E(I)$. The difference is that the multi-layer PD maps to the corresponding private classes. 
As noted before, the architecture and training algorithm of \systemname can easily incorporate other architectures as the privacy discriminator (PD).
The error between the predicted private class $z'$ and the privacy label $z$ measures the specified privacy leak risk $C_{p2}$.

\item \textbf{Intuitive Privacy Reconstructor (PR)} is a usual Encoder turned upside down, composed of multiple un-pooling layers and deconvolutional layers.
The un-pooling operation is realized by feature resizing or nearest-value padding~\cite{bib:mahendran2015:CVPR}.
And then the Deconvolution layer densifies the sparse activation obtained by un-pooling through reverse convolution operations~\cite{bib:zeiler2010:deconvolutional}.
The PR simulates a malicious party and quantifies the intuitive privacy error $C_{p1}$. After obtaining a (binary) version of the Encoder, a malicious party is free to explore any neural architectures to reconstruct the raw data. 
In this work, we examine multiple reconstructor architectures and select the one with the lowest reconstruction error as the specialist privacy reconstructor.
And we also include an exactly layer-to-layer reversed architecture to mirror the Encoder, to emulate a powerful adversarial reconstructor that knows the internals of the Encoder throughout training.
The reconstruction error, \eg Euclidean distance, between $I$ and $I'$ measures the disclosure risk $C_{p1}$ of agnostic privacy information.

\end{itemize}
\end{itemize}

\subsection{Training Algorithm of PAN}
\label{subsec:algorithm}
% \lin{Please update both the text and Algorithm 1 to only present the final weight updating scheme. } \rev{I have updated both of them to only present v2.}

\begin{algorithm}[t]
 \LinesNumbered
 \KwIn{Dataset $\mathbf{T}$}
 \KwOut{PAN's Weights $\{\theta_E, \theta_{UD}, \theta_{PD}, \theta_{PR} \}$ }
 Initialize $\theta_E, \theta_{UD}, \theta_{PD}, \theta_{PR}$ \;
 \For{$n$ epochs}{
Sample mini-batch $I$ of $m$ samples from $\mathbf{T}$\;
\For{$k$ steps}{
Update $\theta_e$ and $\theta_{UD}$ by gradient ascent with learning rate $l_1$: minimize $C_u $ \;
Update $\theta_{PD}$ by gradient ascent with learning rate $l_2$: minimize $C_{p1}$ \;
Update $\theta_{PR}$ by gradient ascent with learning rate $l_3$: minimize $C_{p2}$ \;
}
Update $\theta_E$ and $\theta_{UD}$ by gradient ascent with learning rate $l_4$: 
       minimize $C_{sum}$ \;
} 
*Note: $n$ and $k$ are two hyper-parameters to synchronize the training of E, UD, PD, and PR parts.
 \caption{Mini-batch stochastic training of privacy adversarial network (PAN)}
 \label{alg_ran}
\end{algorithm}

Our goal with \systemname is to train an Encoder that can produce output that is both useful, \ie leading to high inference accuracy when used for classification tasks, and private, \ie leading to low privacy inference accuracy and high reconstructive error when maliciously processed and reversely engineered by the attacker, respectively.
As we noted in \S\ref{sec:problem}, the utility and privacy objectives can be competing when taken naively.
The key idea of the \systemnameposs training algorithm is to train the Encoder along with the utility discriminator and the two types of privacy attackers, which specialize in discrimination and reconstruction, respectively.
%service provider and a malicious attacker, respectively. 
%
Given a training dataset $\mathbf{T}$ of $m$ pairs of $I$, the raw data, $y$, the true task label, and $z$, the privacy label, we train a \systemname through an iterative process with the following four stages:
\begin{enumerate}

\item Discriminative training mainly maximizes the accuracy to train a specialist utility discriminator (UD); mathematically, it minimizes the cross entropy $\tau$ between predicted class $\mathop{UD}(\mathop{E}(I_i))$ and true label $y_i$:
\begin{equation}
\label{equ_cu}
Min\ C_u=\sum_{i=1}^{m} {\tau(y_i, \mathop{UD}(\mathop{E}(I_i))) }.
\end{equation}

\item Discriminative training minimizes the cross entropy $\tau$ between predicted private class $\mathop{PD}(\mathop{E}(I_i))$ and private ground truth $z_i$, to primarily train a specialist privacy discriminator (PD):
\begin{equation}
\label{equ_cp1}
Min \ C_{p1}=\sum_{i=1}^{m} {\tau(z_i, \mathop{PD}(\mathop{E}(I_i))) }
\end{equation}

\item Generative training minimizes the reconstructive error to train a specialist privacy reconstructor (PR):
\begin{equation}
\label{equ_cp2}
Min \ C_{p2}=\sum_{i=1}^{m} {|I_i- \mathop{PR}(\mathop{E}(I_i)) |^2 }
\end{equation}

\item Adversarial training minimizes the sum error to find a \emph{privacy-utility tradeoff}. Specifically, it trains the Encoder to suppress utility error $C_u$ and increase privacy error ($C_{p1}, C_{p2}$):
\begin{equation}
\label{equ_csum}
Min\ C_{sum}= \sum_{i=1}^{m} { \lambda_1 \tau (y_i, \mathop{UD}(\mathop{E}(I_i))) -  \lambda_2 |I_i- \mathop{PR}(\mathop{E}(I_i))|^2 - \lambda_3 \tau (z_i, \mathop{PD}(\mathop{E}(I_i)))  } 
\end{equation}
$C_{sum}$ is a Lagrangian function of $C_u, C_{p_1}$ and $C_{p2}$.  $\lambda_1$, $\lambda_2$, and $\lambda_3$ are Lagrange multipliers that can be used to \lsc{control the relative importance of privacy and utility}.
When we set $\lambda_2=0$ or $\lambda_3=0$, \systemname only trains the Encoder to resist against the specified privacy discriminator or the intuitive privacy reconstructor, respectively. 
%\lsc{The value of $C_{sum}$ is regarded as a quantitative metric of utility-privacy tradeoff  in $\S$~\ref{sec:experiment}.}
\end{enumerate}

Algorithm \ref{alg_ran} summarizes the training algorithm of \systemname. 
We leverage mini-batch techniques to split the training data into small batches, over which we calculate the average of the gradient to reduce the variance of gradients, which
balance the training robustness and efficiency (line 3)~\cite{bib:li2014:sigkdd}.
Within each epoch, we first perform the standard discriminative and generative stages (line 5, 6, 7) to initialize the Encoder's weights $\theta_E$ and train the specialist utility discriminator (UD), privacy discriminator (PD) and privacy reconstructor (PR).
And then, we perform the adversarial stage (line 9) to shift the utility-privacy tradeoff on the Encoder weight $\theta_E$ tuning.
We note that $k$ in line 4 is a hyper-parameter of the first three stages. These $k$ steps followed by a single iteration of the forth stage is trying to synchronize the convergence speed of these four training stages well, borrowing existing techniques in generative adversarial network~\cite{bib:goodfellow2014:advances}.
Our implementation uses an empirically optimized value of $k=3$.
And we leverage the AdamOptimizer~\cite{bib:kingma2014:arxiv} with an adaptive learning rate for all four stages (line 5, 6, 7 and 9).

%% file: body/experiment.tex
\section{Evaluation}
\label{sec:experiment}

In this section, we evaluate \systemnameposs performance using six classification services for mobile apps, with a focus on the utility-privacy tradeoff. We compare \systemname against alternative methods reported in the literature and visualize the results for insight into why \systemname excels.

\begin{table*}[t]
\centering
\scriptsize
\caption{Summary of the mobile applications and corresponding datasets for evaluating \systemname.}
  \label{tb:task}
 % \vspace{-4mm}
  \begin{tabular}{|l|l|l|l|l|}
  \hline
  \textbf{No.} & \textbf{\lsc{Target task (utility label)}} & \textbf{\lsc{Private attribute (privacy label)}} & \textbf{Dataset} & \textbf{Description} \\ \hline
  $T_{1}$ & Digit ($10$ classes) & None & MNIST~\cite{data:mnist1998:LeCun} & $70,000$ images 
  %($60,000$ for training and $10,000$ for testing), 
  \\ \hline
  $T_{2}$ & Image ($10$ classes) & None & CIFAR-10~\cite{data:cifar}  & $60,000$ images
  %($50,000$ for training and $10,000$ for testing),
   \\ \hline
  $T_{3}$ & Image ($5$ classes) & None & ImageNet~\cite{data:imagenet} & $65,000$ images
  %($60,000$ for training and $5,000$ for testing),
   \\ \hline
  $T_{4}$ & Acoustic event ($9$ classes) & None & UbiSound~\cite{bib:sicong2017:IMWUT}& $7,500$ audio clips 
  %($6,500$ for training and $1,000$ for testing),
   \\ \hline
  $T_{5}$ & Human activity ($7$ classes) &Human identity ($33$ classes) & Har~\cite{data:Har} & $10,000$ records of accelerometer and gyroscope
  %($7,300$ for training and $2,700$ for testing),
    \\ \hline
  $T_{6}$ & Driver behavior ($10 $ classes) & Driver idenity ($ 26$ classes) & StateFarm~\cite{data:statefarm} & $22,424$ images
  %($7,300$ for training and $2,700$ for testing),
   \\ \hline
\end{tabular}
%\vspace{-4mm}
\end{table*}

\subsection{Experiment Setup}
%We first present the settings for our evaluation.

\textbf{Evaluation applications $\&$ datasets.}
We evaluate \systemname, especially the resulting Encoder, with six commonly used mobile applications/services, for which the corresponding benchmark datasets are summarized in~\tabref{tb:task}. 
\lsc{
Specifically, the target task in $T_{1}$ (MNIST ~\cite{data:mnist1998:LeCun}) is handwritten digit recognition. 
The agnostic private information in the real-world raw image may include individual handwritten style and the background paper.
We use $50,000$ images for \systemname training and $20,000$ images for validation and testing.
The target tasks in $T_{2}$ (CIFAR-10~\cite{data:cifar}) and $T_{3}$ (ImageNet~\cite{data:imagenet}) are image classification. 
The agnostic private information in the real-world raw image may involve background location, color, and brand.
We choose $40,000$ images for training and remaining images for testing in both cases.
The target task in $T_{4}$ (UbiSound~\cite{bib:sicong2017:IMWUT}) is to recognize acoustic event. 
The agnostic private information covers background voice and environment information.
We use $6,000$ audio clips for training and $1,500$ audio clips for testing.
The target task in $T_{5}$ (Har~\cite{data:Har}) is human activity identification based on the records of accelerometer and gyroscope. The specified private attribute we intend to hide is useridentity. 
And the agnostic private information we expect to protect may contain individual habit.
We randomly select $8,000$ records for training and $2,000$ records for testing.
The target task in $T_6$ (StateFarm~\cite{data:statefarm}) is to predict driver behavior. The specified private attribute we choose is driver identity. 
And the agnostic private information within the real-world raw image can be face and gender.
We use $18,000$ images for training and $4,424$ images for testing.
}
% \systemname is evaluated in terms of \lsc{\emph{task utility}} by the inference accuracy on 
%
% \systemname is tested for the \emph{specified privacy}, \ie inference accuracy on private attribute: human identity ($T_{5}$: Har~\cite{data:Har}) and driver identity ($T_6$: StateFarm~\cite{data:statefarm}).
% inference (\emph{specified privacy}) with human identity label , and images based driver behavior recognition with driver identity label .
%

\textbf{Evaluation models.} In \systemname, we leverage a utility discriminator (UD), a privacy discriminator (PD), and a privacy reconstructor (PR) model to train and validate the Encoder (E).
In the training phase, we refer to the successful neural network architectures to build  \systemnameposs Encoder (E), Utility Discriminator (UD) and Privacy Discriminator (PD) for different types of datasets.
For example, according to the sample shape in the datasets, the LeNet is chosen as the reference for $T_1$, $T_4$ and $T_5$, AlexNet is for $T_2$ and $T_6$, and VGG-16 model is for $T_3$.
To evaluate the learned Encoder in the testing phase, we leverage another set of separately trained Utility Discriminator (UD) and Privacy Attackers (PD and PR), given \systemnameposs Encoder output, to simulate the service provider and malicious parties.
In particular, we ensemble multiple optional MLP architectures to simulate the service provider's Utility Discriminator (UD) for task recognition, as well as the malicious attacker's Privacy Discriminator (PD) for private attribute prediction.
These MLP models have different fully-connected architectures by using varying scales of  singular value decomposition, sparse-coding factorization, and global-average computation to replace the initial fully-connected layers. 
We also employ multiple generative architectures to select the most powerful one as the privacy reconstruction attacker (PR).
To emulate a powerful adversary that knows the Encoder for the attackers' training, we include a PR model that exactly mirrors the Encoder for each task.

\textbf{Prototype implementation.} \lsc{\systemname has two phases: an offline phase to train the Encoder, and an online phase where we deploy the learned Encoder as a middleware on mobile platforms to encode the raw sensor data into features.}
In the offline phase, we use the Python library of TensorFlow~\cite{lib:tf} to train the Encoder, utility discriminator, privacy discriminator as well as privacy reconstructor \lsc{using the datasets summarized in~\tabref{tb:task}}. And we leverage h5py ~\cite{lib:h5py} to separately save the trained models. To speedup the training, we leverage a server with four Geforce GTX 1080 Ti GPUs with CUDA 9.0.
% while deploying the trained Encoder locally on the mobile platform as an effective and compact mobile privacy-preserving tool.
%
\lsc{In the online phase, we prototype the mobile-side on the Android platform,} \ie Xiaomi Mi6 smartphone, using TensorFlow Mobile Framework~\cite{lib:tf:mobile}. And we store the learned Encoder in the smartphone's L2-cache using Android's LuCache API~\cite{bib:LruCache}, \lsc{which speeds up the on-device data encoding}.
\lsc{
The Encoder intercepts the incoming testing data and encodes it into features, which are then fed into the corresponding Android Apps for real-word performance evaluation.}

\begin{figure*}[t]
  \centering
  \subfloat[MNIST ($T_1$)]{
  \includegraphics[width=0.26\textwidth]{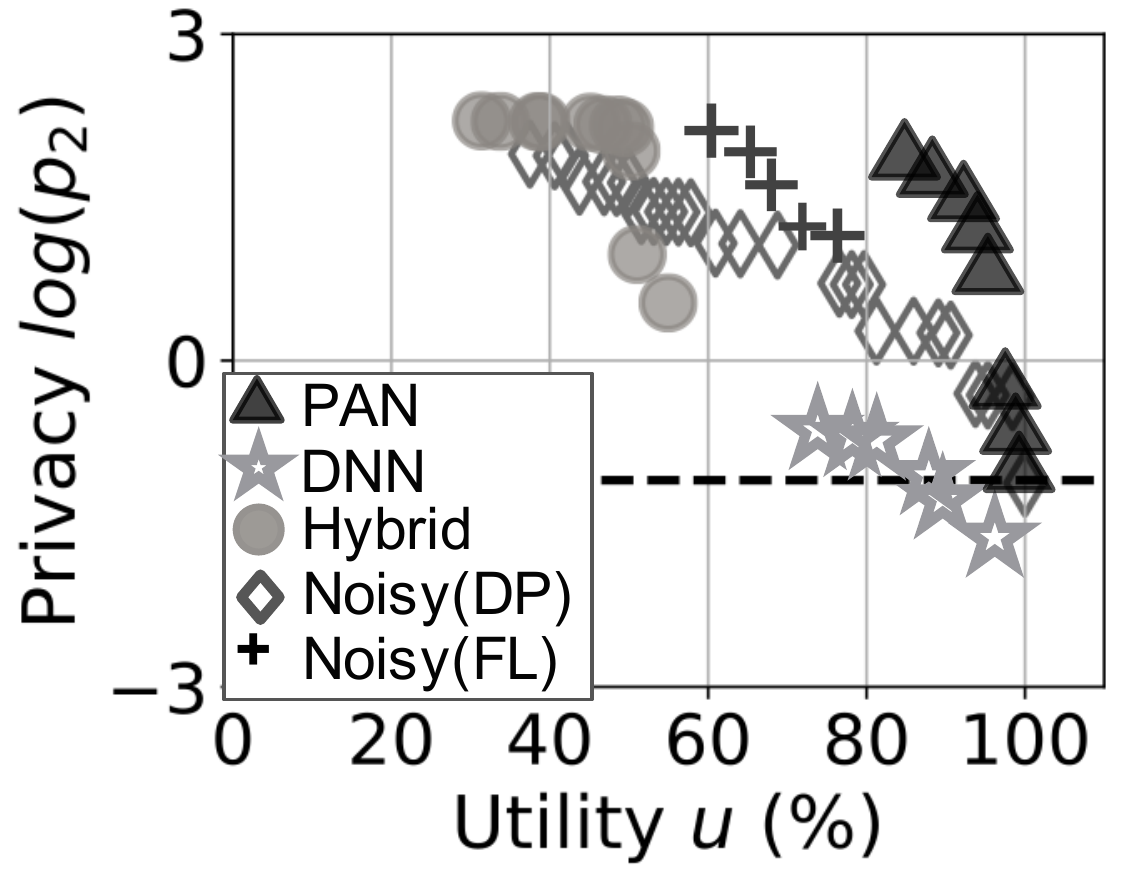}}
  \hfill
    \subfloat[CIFAR-10 ($T_2$)]{
  \includegraphics[width=0.26\textwidth]{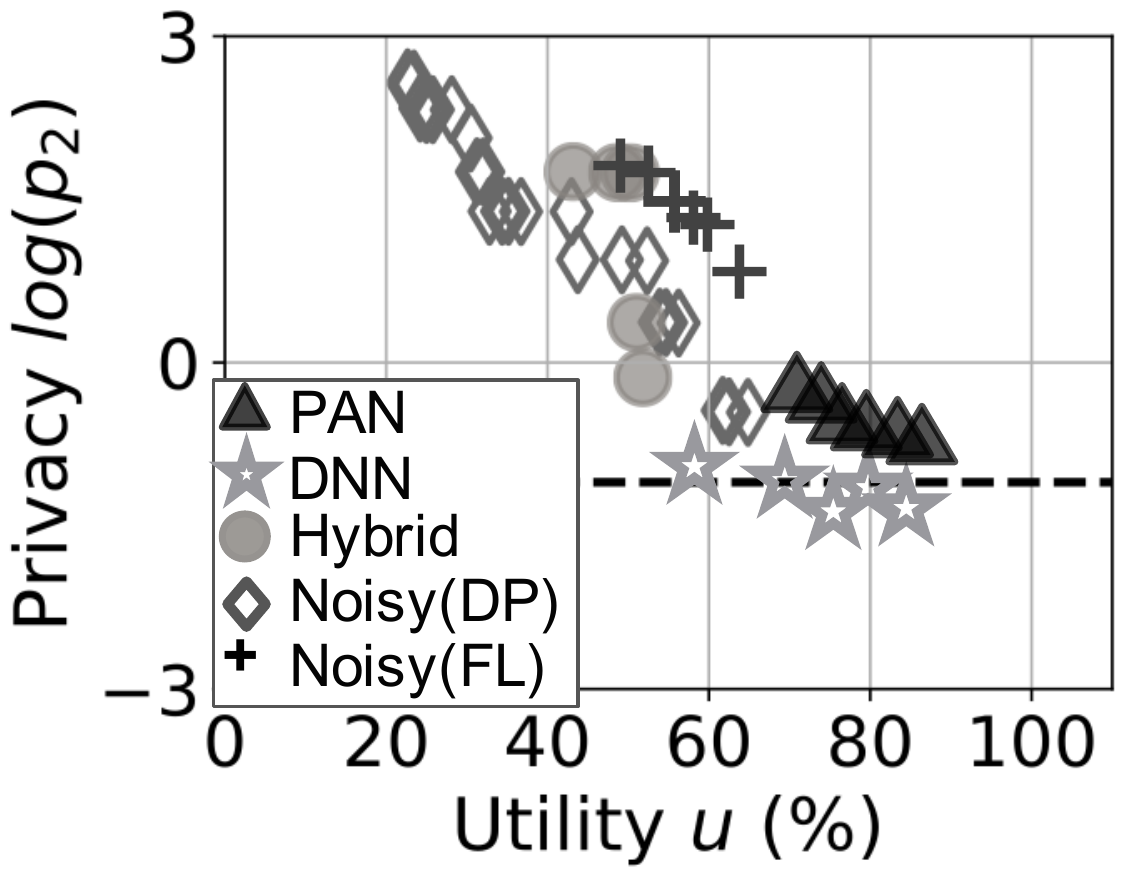}}
    \hfill
    \subfloat[ImageNet ($T_3$)]{
  \includegraphics[width=0.26\textwidth]{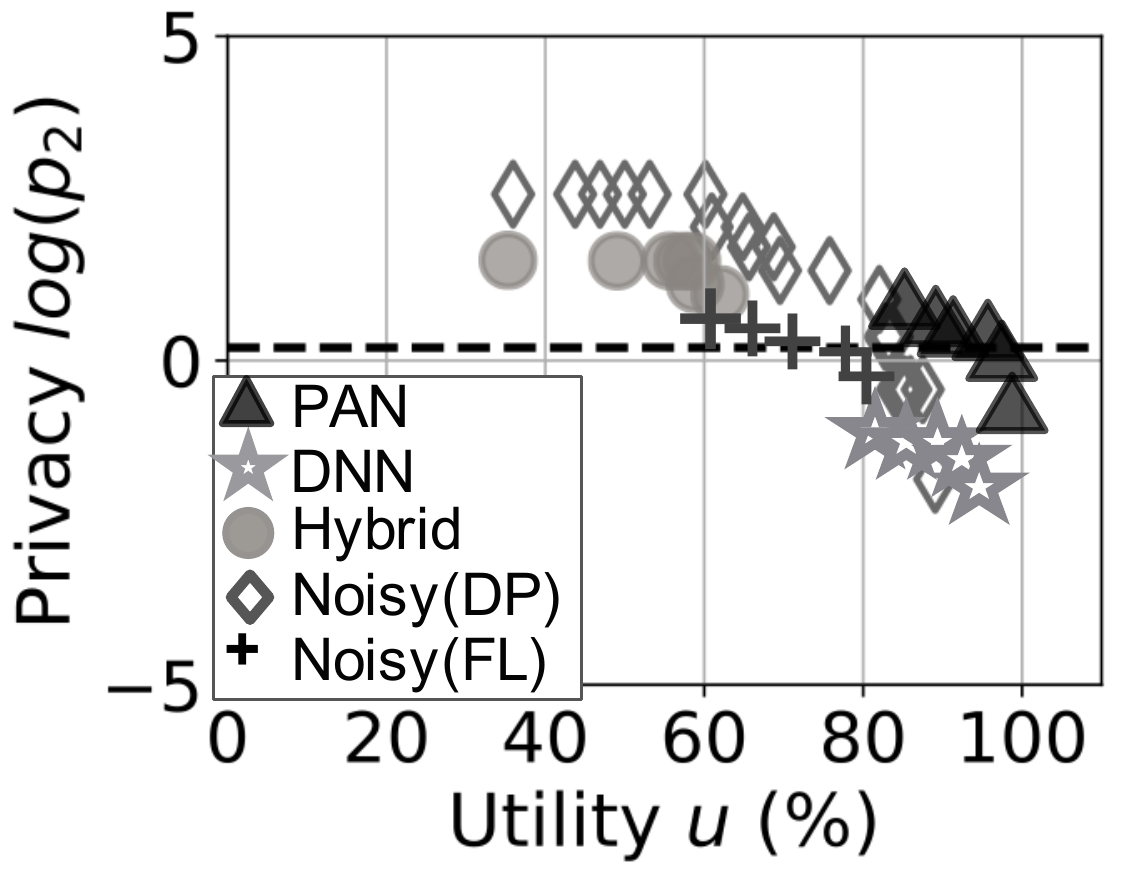}}
  \hfill
      \subfloat[UbiSound ($T_4$)]{
  \includegraphics[width=0.26\textwidth]{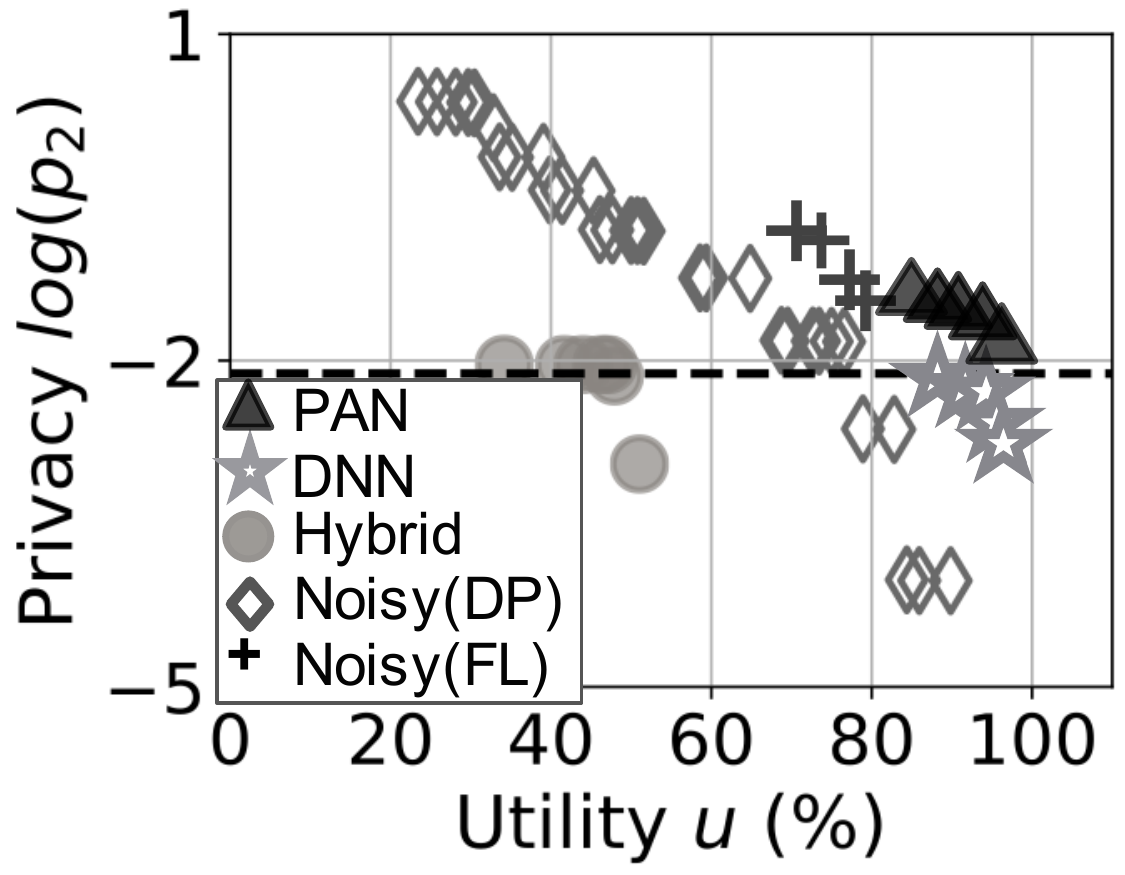}}
    \hfill
      \subfloat[Har ($T_5$)]{
  \includegraphics[width=0.26\textwidth]{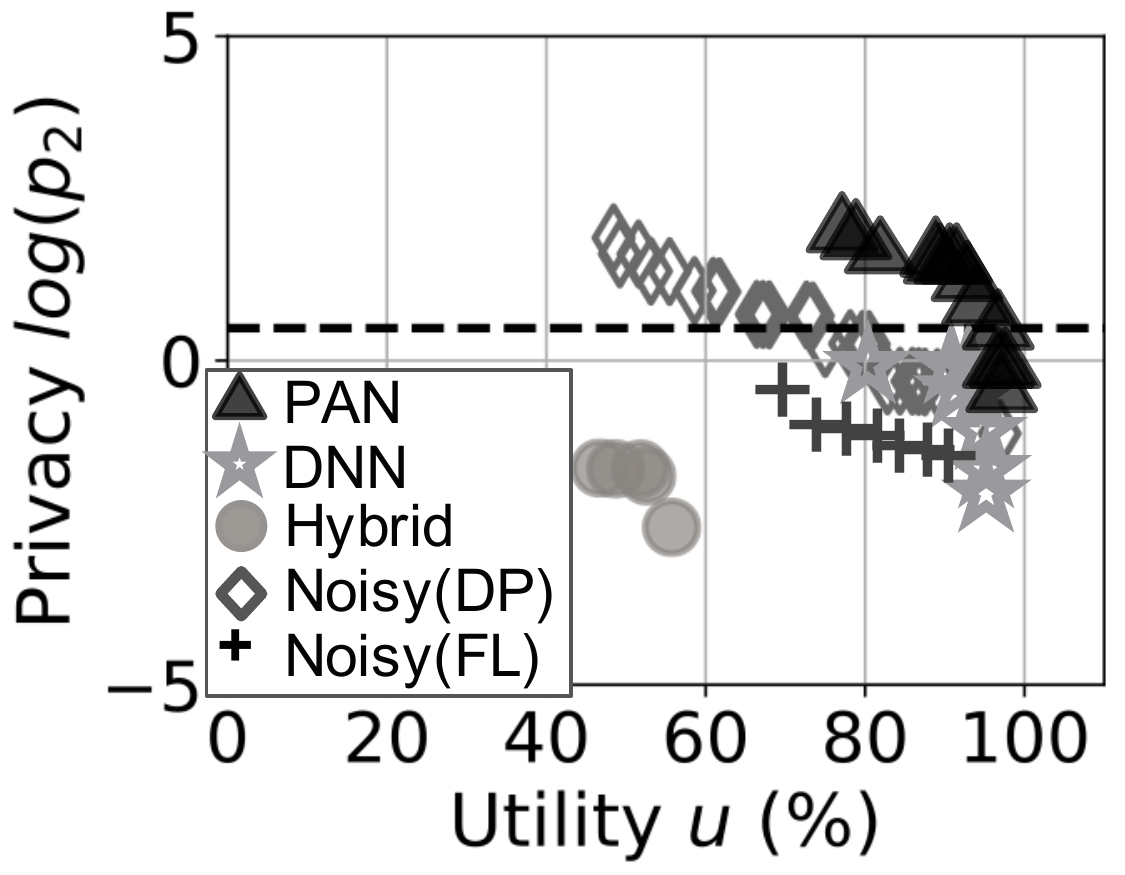}}
    \hfill
      \subfloat[StateFarm ($T_6$)]{
  \includegraphics[width=0.26\textwidth]{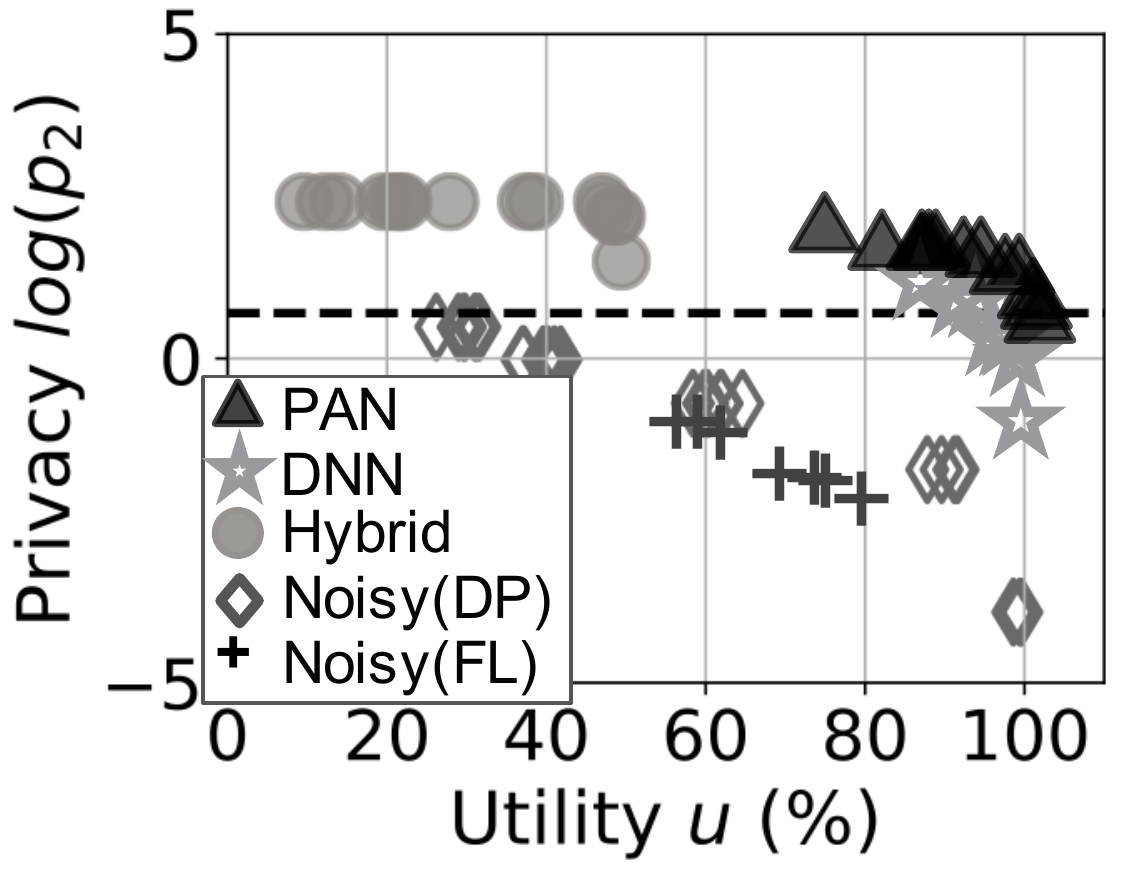}}
%  \vspace{-4mm}
\caption{Performance comparison of PAN$_1$ with four baselines across six different applications ($T_1, T_2, T_3, T_4, \lsc{T_5, T_6}$).
The X-axis shows the utility $u$ (\ie task inference accuracy) tested by the simulated service provider. And Y-axis is the intuitive reconstruction privacy $p_2$ tested by the simulated malicious attacker, normalized by $log$ operation. \lsc{The dashed line sets a privacy $log(p_2)$ benchmark validated by \systemnameposs privacy reconstructor}.}
\label{fig_compare_pan1}
%\vspace{-6mm}
\end{figure*}

\subsection{Comparison Baselines}
\label{subsec:baseline}
We employ \lsc{four} types of state-of-the-art data privacy preserving baselines to evaluate \systemname. The DNN method provides a high utility standard, and the DP, FL, and Hybrid DNN methods set a strict utility-privacy tradeoff benchmark for \systemname. The detail settings of the baseline approaches and \systemname are as below.
\begin{itemize}[leftmargin=*]
\item \textbf{\lsc{Noisy (DP)}} method perturbs the raw data $I$ by adding Laplace noise with diverse factors $\{0.1, 0.2, ... 0.9\}$, and then submit the noisy data $\overline{I}$ to the service provider. This is a typical local differential privacy (DP) method~\cite{bib:he2017:ACC, bib:dwork2010:STC}.
% As shown in Fig.~\ref{fig_baseline}, 
%
The utility $u$ of noisy data is tested by the task \lsc{(\eg the driver behavior in $T_6$)} recognition accuracy in a MLP classifier $UC$ with multiple fully-connected layers.
%, \lsc{ \ie $u=prob(UC(I_i)=y_i), I_i \in T_{test}$. Here $prob$ denotes the probability of true prediction, $y_i$ is the utility label, and $T_{test}$ is the corresponding testing set within datasets $T_1 \sim T_6$.}
%which are trained on noisy data. 
%
The specified privacy $p_1$ \lsc{is measured by the inference accuracy over private attribute (\eg driver identity in $T_6$)} in another MLP classifier $PC$.
% , \lsc{\ie $p_1=prob(PC(I_i)=z_i), I_i \in T_{test}$. Here, $z_i$ is the privacy label (see \tabref{tb:task}).}
%
And the intuitive privacy $p_2$ is evaluated by the average information loss, \ie $p_2=avg(|I_i-\overline{I_i}|^2), I_i \in T_{test}$. Here $T_{test}$ is the corresponding testing set within datasets $T_1 \sim T_6$.
%
% To simulate user's randomness to add noise to the raw data, 
% We add Laplace noise with diverse noise factors to train DNN and test the utility and privacy.

% ($0.1, 0.2, 0.3, 0.4, 0.5, 0.6, 0.7, 0.8, 0.9$) to test the performance of noisy data method.
% \lin{``Laplace noise'' was never properly introduced. What is its intellectual context? Differential privacy?}
% \rev{Great advice! as I checked, it is widely used in differential privacy.}

\item \textbf{\lsc{Noisy (FL)}} \lsc{method perturbs the data $I$ by adding Gaussian noise $N(0, \sigma^2)$ with mean $0$ and variance $\sigma^2$, where we set $\sigma=40$ according to ~\cite{bib:papernot:ICLR2018}. The Gaussian noise included in the noisy data $\overline{I}$ can provide rigorous guarantees of differential privacy using less local noise. This is widely used in the noisy aggregation scheme of federated learning (FL)~\cite{bib:truex:arXiv2018, bib:papernot:ICLR2018}.
We test the utility $u$, the specified privacy $p_1$, and the intuitive privacy $p_2$ of this noisy data $\overline{I}$ using the similar methodology as DP baseline.
}
\item \textbf{DNN} method encodes the raw data $I$ into features $F$ using a deep encoder with multiple convolutional and pooling layers, and expose features $F$ to the service provider~\cite{url:googlecloud, url:googlenow}. 
The utility $u$ of DNN features is measured by the inference accuracy in a classifier $UC$ with multiple fully-connected layers.
% , \ie $u=prob(UC(F_i)=y_i), I_i \in T_{test}$.}
%
The specified privacy $p_1$ is tested by the inference accuracy over the private attribute in another privacy classifier $PC$ with multiple fully-connected layers.
% , \lsc{\ie $p_1=prob(PC(F_i)=z_i), I_i \in T_{test}$.}
%
And the intuitive privacy $p_2$ is tested by the reconstruction error in \lsc{a decoder $D$ with multiple deconvolutional and unpooling layers, \ie $p_2=|I_i-D(F_i)|^2, I_i \in T_{test}$.}

%and the the accuracy evaluates the utility in a DNN based classifier (\eg the fully-connected layers of LeNet, AlexNet, VGG).

% \lin{DNN or CNN? Be consistent. Also, to be consistent with ``RAN'', I would call this ``DNN'', instead of ``Feature''. The next one ``DNN (resized)'' instead of ``Resized Feature''}

\item \textbf{Hybrid} DNN method
% \lin{Can we cite more papers and practices, e.g., Google Now? The point is to highlight that we didn't pick this baseline randomly.}
% As shown in Fig.~\ref{fig_baseline2}, 
further perturbs the above DNN features through additional lossy processes, \ie principal components analysis (PCA) and adding Laplace noise~\cite{bib:ossia17:arxiv} with varying noise factors $\{0.1, 0.2, ... 0.9\}$, before delivering them to the service provider. 
The utility $u$, the specified privacy $p_1$, and the intuitive privacy $p_2$ of the perturbed features $F'$ is respectively tested by a task classifier, a private attribute classifier, and a decoder, using the same methodology of the DNN baseline.  

\item \textbf{PAN} automatically transform raw  data $I$ into features, \ie $E(I)$, using the learned Encoder $E(\cdot)$.
% a utility discriminator (UD) for utility $C_u$ testing, to a privacy discirminator (PD) for specified privacy $C_{p1}$ testing, and to a privacy reconstructor (PR) for intuitive  privacy $C_{p2}$ evaluation, the same as the DNN baseline.
%
In particular, we evaluate the following two types of \systemname, that are trained \lsc{to defend against different types of privacy attackers for different benchmark tasks/datasets (\tabref{tb:task})}:
\begin{itemize}[itemindent=1em]
\item \textbf{PAN$_1$} is trained with one privacy attacker, \ie the Privacy Reconstructor (PR), by setting $\lambda_3 =0$ in the adversarial training objective (\equref{equ_csum}). We train the PAN$_1$ on six datasets (\ie $T_1 \sim T_6$).
\item \textbf{PAN$_2$} is trained with two privacy attackers, \ie Privacy Discriminator (PD) and Privacy Reconstructor (PR), for application datasets accompanied with both utility labels and private attribute labels (\ie $T_5$ and $T_6$).
\end{itemize}
The utility $u$ of both $PAN_1$'s and $PAN_2$'s Encoder output are tested by the task inference accuracy in the service provider's utility discriminator (UD) using a classifier.
% , \lsc{\ie $u=prob(UD(E(I_i))=y_i), I_i \in T_{test}$}.
%
The specified privacy $p_1$ of $PAN_2$'s Encoder output is evaluated by the inference accuracy in the attacker's privacy discriminator (PD) using a classifier.
% , \lsc{\ie $p_1=prob(PD(E(I_i))=z_i), I_i \in T_{test}$}.
%
As for the intuitive reconstruction privacy $p_2$ in PAN$_1$ and PAN$_2$, we select the most powerful decoder as the privacy reconstructor (PR) to evaluate it, \lsc{\ie $p_2=|I_i-PR(E(I_i))|^2, I_i \in T_{test}$}.
% simulate a powerful malicious attacker.
\end{itemize}

\begin{figure*}[t]
  \centering
  \subfloat[$p_1$, Har ($T_5$)]{
  \includegraphics[width=0.24\textwidth]{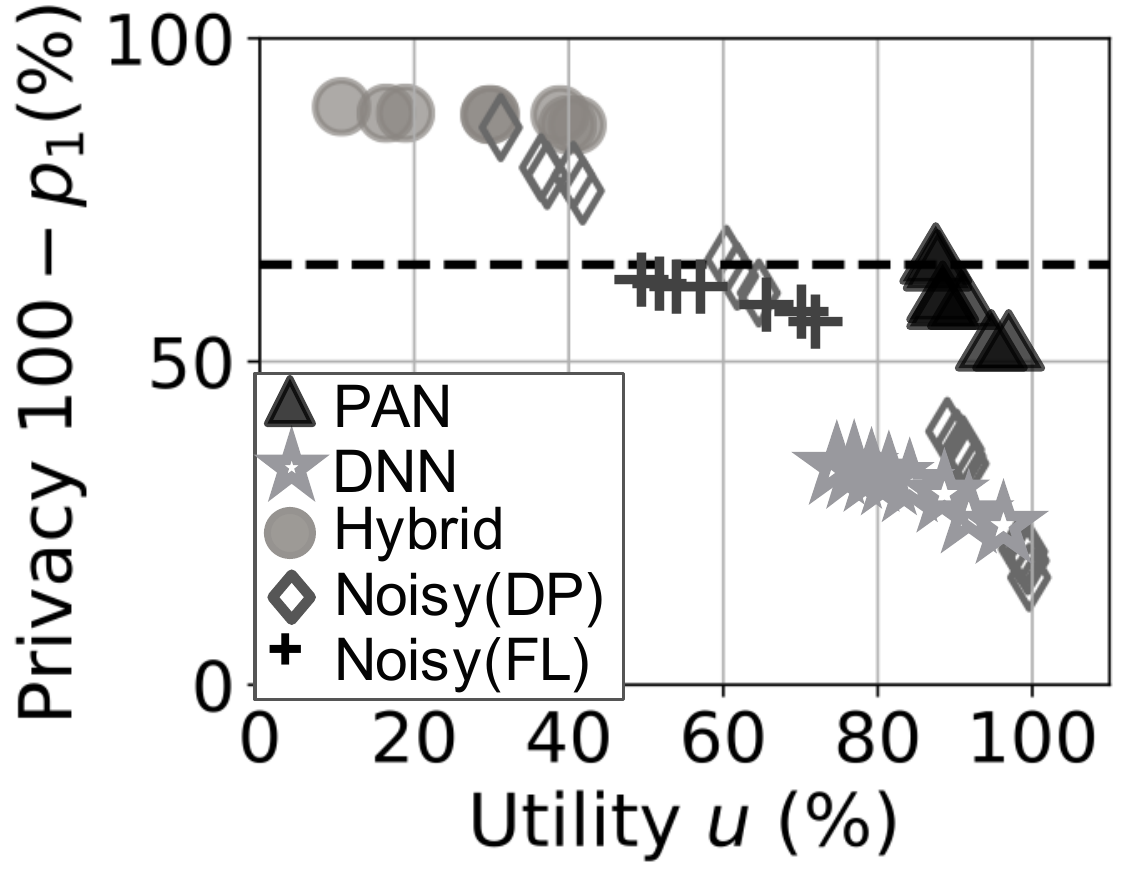}}
   \hfill
\subfloat[$p_2$, Har ($T_5$)]{
  \includegraphics[width=0.24\textwidth]{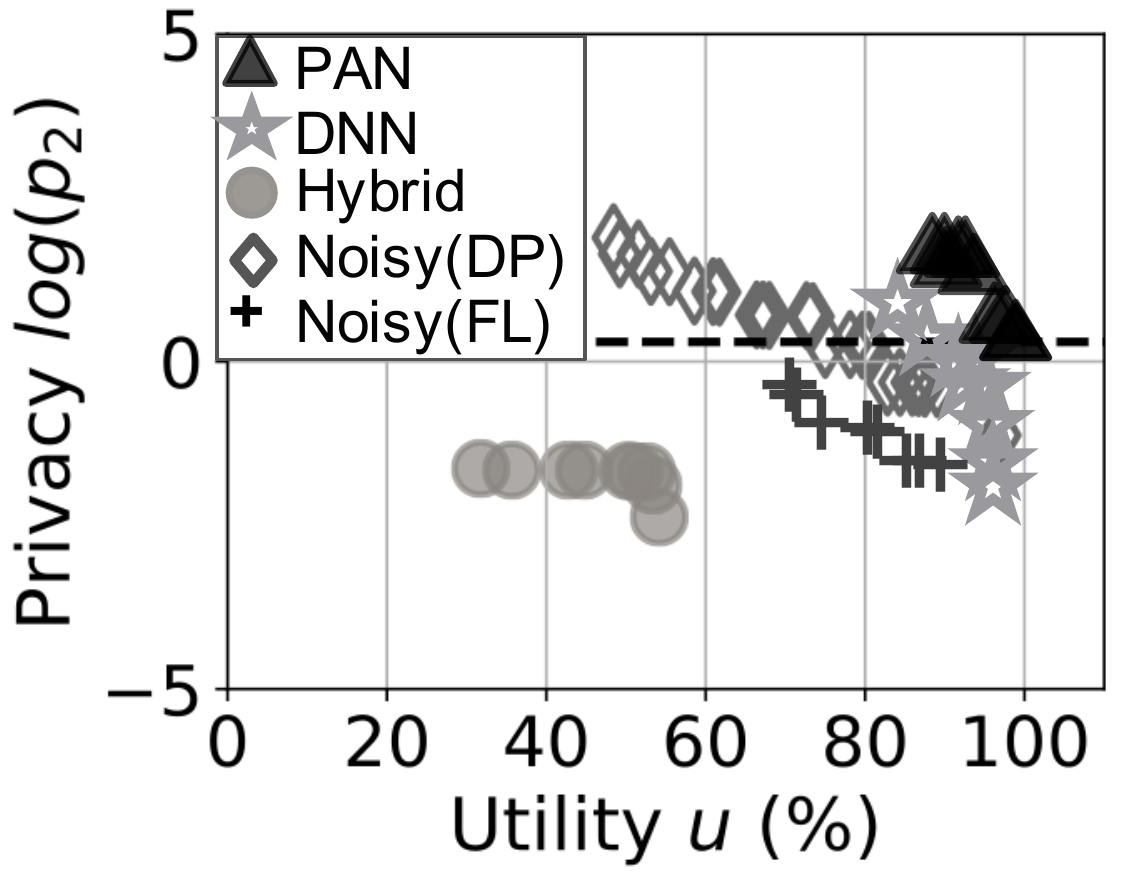}}
 \hfill
  \subfloat[$p_1$, StateFarm ($T_6$)]{
  \includegraphics[width=0.24\textwidth]{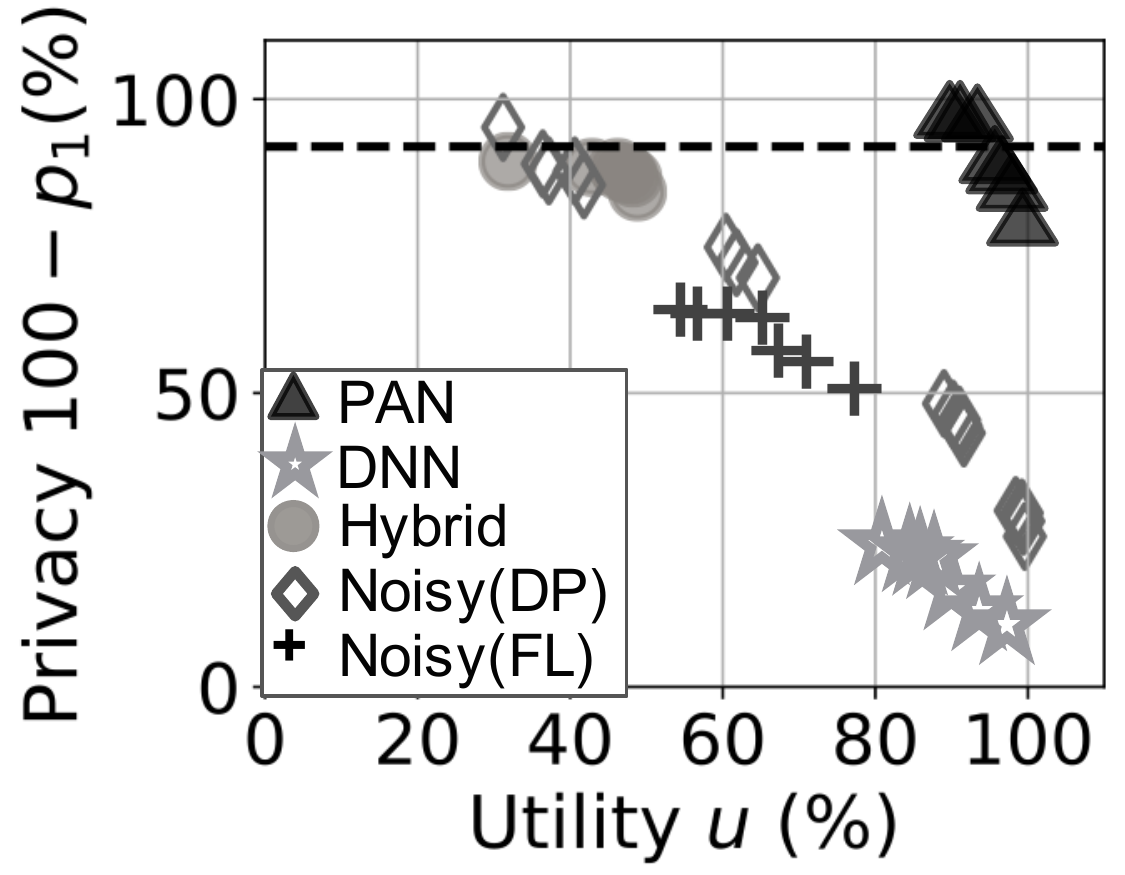}}
 \hfill
    \subfloat[$p_2$, StateFarm ($T_6$)]{
  \includegraphics[width=0.24\textwidth]{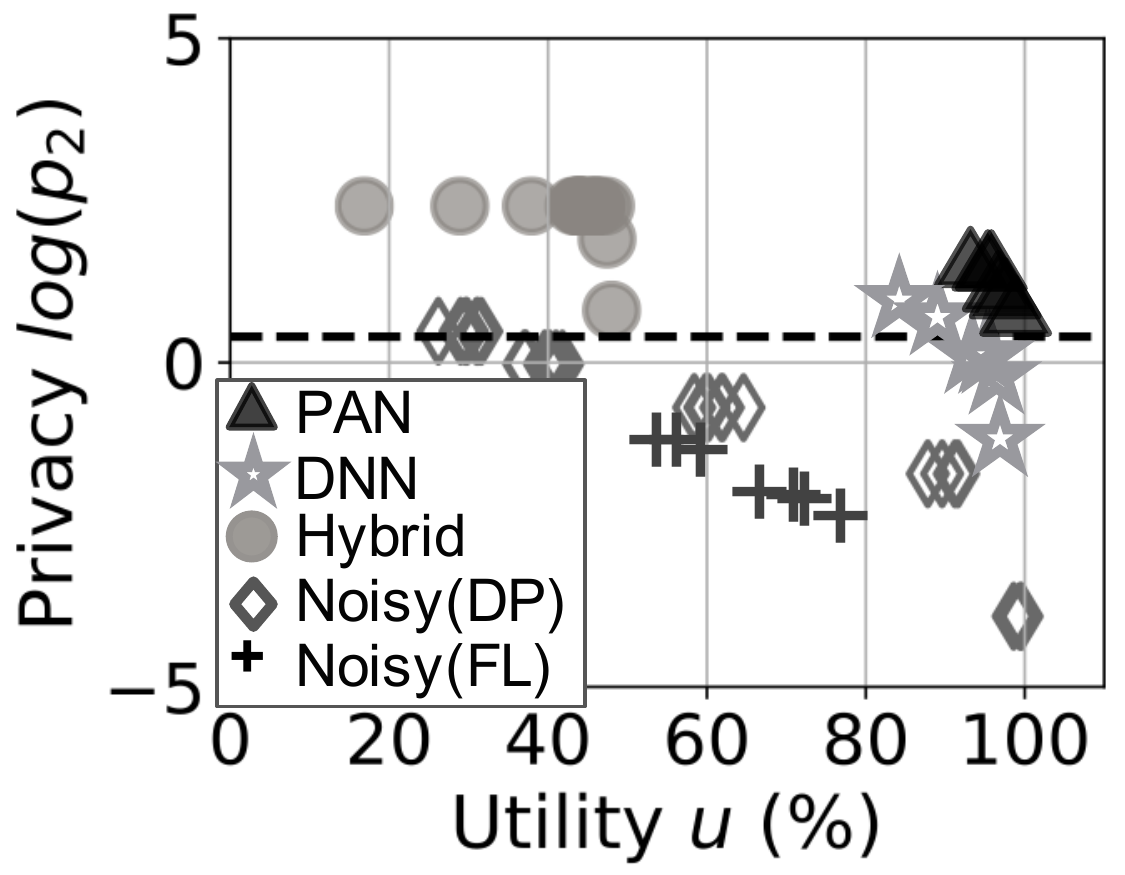}}
%  \vspace{-4mm}
\caption{Comparison of PAN$_2$ with four baselines on \lsc{Har ($T_5$)} and StateFarm ($T_6$).
X-axis is the utility $u$ (\ie task inference accuracy). \lsc{Y-axis in (a) and (c) represents the specified privacy $p_1$ (\ie private attribute inference accuracy) normalized by $100-p_1(\%)$. And Y-axis in (b) and (d) is the intuitive  privacy $p_2$ normalized by $log$ operation}. The dashed line set the privacy benchmark validated by \systemnameposs privacy discriminator in (a) and (c) and privacy reconstructor in (b) and (d).}
\label{fig_compare_state}
%\vspace{-4mm}
\end{figure*}

\subsection{Utility vs. Privacy Tradeoffs}
\label{subsec:exp_tradeoff}
This subsection evaluates \systemname in terms of the utility $u$ by the service provider and the privacy $p_1$ and $p_2$ by the malicious attackers, compared with \lsc{four} privacy-preserving baselines (see $\S$ \ref{subsec:baseline}).
\figref{fig_compare_pan1} and \figref{fig_compare_state} summarize the Pareto fronts of the testing privacy-utility tradeoffs by \lsc{four} baselines and \systemname.
In this set of experiments, we train the PAN$_1$ on the six application datasets ($T_1 \sim T_6$) \lsc{based on utility labels}, and train the PAN$_2$ on Har ($T_5$) and StateFarm ($T_6$) datasets \lsc{accompanied with both utility labels and private attribute labels (see Table~\ref{tb:task})}. 

First, \systemnameposs Encoder output achieves the best privacy-utility tradeoff, compared to those encoded by other four baselines.
\lsc{In \figref{fig_compare_pan1}, we see the performance of PAN$_1$'s Encoder output lies in the upper right corner with maximized utility $u$ and maximized privacy $log(p_2)$ on the digit recognition applications ($T_1$), and lie around the upper right side with maximized utility $u$ and competitive privacy $log(p_2)$ compared with other four baselines on image classification applications ($T_2$ and $T_3$) and audio sensing applications ($T_4$).}
In \figref{fig_compare_state}, the PAN$_2$ \lsc{is also in the upper right corner with maximized utility $u$ and maximized privacy $(100-p_1)\%$ or $log(p_2)$} on both human activity recognition application ($T_5$) and driver behavior prediction application ($T_6$).
\lsc{Here we transform the expected minimized privacy $p_1\%$ to the maximized privacy $(100-p_1)\%$}.
\lsc{When we consider the $\lambda_1 u + \lambda_2 (100-p_1) - \lambda_3 log(p_2)$ as a quantifiable metric of utility-privacy tradeoff, both the PAN$_1$'s and PAN$_2$'s Encoder output achieve the best tradeoff value according to the default relative importance $\lambda_1=0.4, \lambda_2=0.3$ and $\lambda_3=0.3$.
While the DNN method provides unacceptable low privacy, and Hybrid DNN, Noisy (DP) and Noisy (FL) methods offer high privacy at the cost of utility degradation.}
\lsc{
Second, the utility (\ie task inference accuracy) of \systemnameposs Encoder output is at least as good as and sometimes even better than the other four baseline methods across different applications.
Specifically, the task inference accuracy by PAN$_1$ is $83.1 \sim 99.8\%$ on MNIST ($T_1$), $70.5 \sim 89.3\%$ on CIFAR-10 ($T_2$), $81.8 \sim 99.2\%$ on ImageNet ($T_3$), $83.6 \sim 98.1\%$ on UbiSound ($T_4$), $78.6 \sim 98.5\%$ on Har ($T_5$), and $77.8 \sim 98.6\%$ on StateFarm ($T_6$), maintaining at a high level.
And the task inference accuracy by PAN$_2$ is $85.1 \sim 99.3\%$ on Har ($T_5$) and $87.3 \sim 99.8 \%$ on StateFarm ($T_6$).
It is even better than the utility of standard DNN features.
% with the proper $\lambda$ setups, which is even larger than that of the original deep model (see DNN baseline).
%
\lsc{Although, with carefully-calibrated Gaussian noise distribution in Noisy (FL), we observe better utility when using Noisy (FL) method than that using the Noisy (DP) method.}
The task inference accuracy in Noisy (DP), Noisy (FL) and Hybrid DNN baselines is seriously unstable, ranging from $12.8\%$ to $96.7\%$ on different applications, because of the injected noises.}
Also, we see the utility of PAN$_2$ on Har and StateFarm is slightly improved than the PAN$_1$ case. It implies the PAN$_2$ with two adversaries learns better features than PAN$_1$ with only one generative model-based adversary.
Third, \systemnameposs Encoder output in both PAN$_1$ and PAN$_2$ cases considerably improves the privacy than the DNN method and achieves the competitive privacy compared with other three baselines.
Moreover, the \systemnameposs privacy $p_1$ and $p_2$ quantified by \systemnameposs privacy discriminator (PD) and privacy reconstructor (PR) (the dashed lines in Figure 3,4) is comparable with that measured by the third-party attackers (solid black triangles in Figure 3,4). 
We train the third-party attackers using the binary version of \systemnameposs Encoder.
This result demonstrates the strong adversary ability of \systemnameposs privacy discriminator and privacy reconstructor.

% An important step for \systemname is to determine the Lagrangian multiplier $\lambda$ in adversarial training stage, \ie Eq.(4), which impact \systemnameposs performance tradeoff.
% 
% As shown in Fig.~\ref{fig_lambda}, the best performance tradeoff comes from different Lagrangian multiplier for diverse tasks.
% % in the optimization objective $O_a$ shown in Eq.(4). 
% %
% We compare seven discrete settings of $\lambda$: $(0.3, 0.4, 0.5, 0.6, 0.7, 0.8, 0.9)$ to evaluate the impact of Lagrangian multiplier on \systemnameposs performance.
% % 
% % The best balance coefficient $\lambda$ with the optimal training tradeoff for diverse tasks varies.
% %
% Particularlly, the optimal Lagrange multiplier for digit task (MNIST) is $\lambda=0.4$, for image (CIFAR-10) is $\lambda=0.5$ , for larger image (ImageNet) is $\lambda=0.6$, for human activity (Har) is $\lambda=0.7$ , and for non-speech sound (Ubisound) is $\lambda=0.8$.

\vspace{1ex}\textbf{Summary.} First, \lsc{although the PAN$_1$ and PAN$_2$ cannot always outperform the baseline methods in both utility and privacy, it achieves the best Pareto front for the utility-privacy tradeoffs across various adversaries and applications}. 
Second, the utility, \ie inference accuracy, of \systemnameposs Encoder output is even better than taht of the standard DNN. We will revisit this surprising result in $\S$~\ref{subsec:visualize} and $\S$~\ref{sec:discuss}. 
% We defer the interpretation about how and why it works to $\S$~\ref{sec:discuss}.
%
% We also note that the regularization parameter $\lambda$ in RAN can be further systematically fine-tuned, e.g., exponentially varied using reinforcement learning, so that discovers a better privacy-utility tradeoff.

%
\begin{figure}[t]
  \centering
  \subfloat[PAN$_1$ on MNIST ($T_1$)]{
  \includegraphics[height=0.195\textwidth]{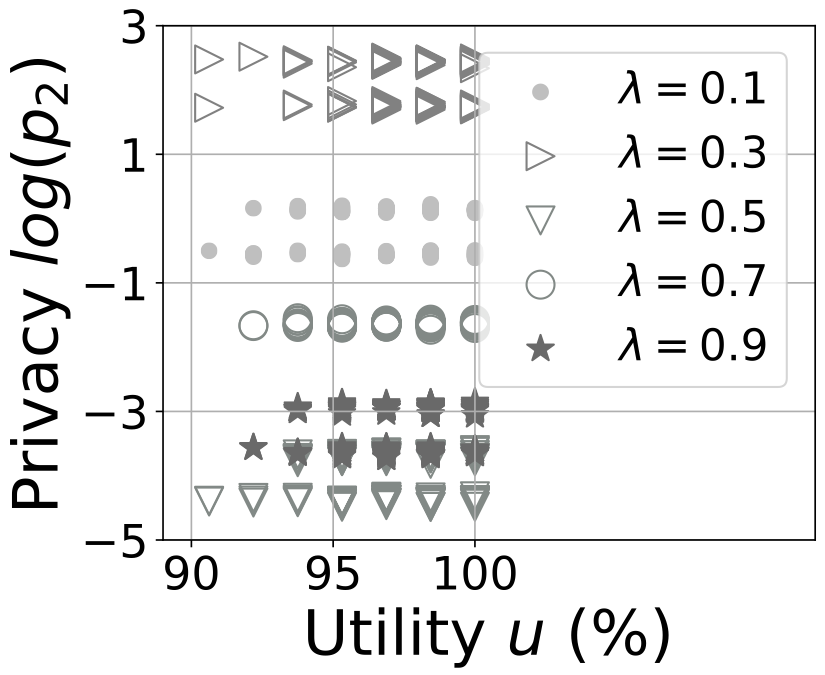}}
%  \subfloat[CIFAR-10]{
%   \includegraphics[width=0.187\textwidth]{image/plot_lambda_cifar}}
   \subfloat[PAN$_1$ on UbiSound ($T_4$)]{
   \includegraphics[height=0.195\textwidth]{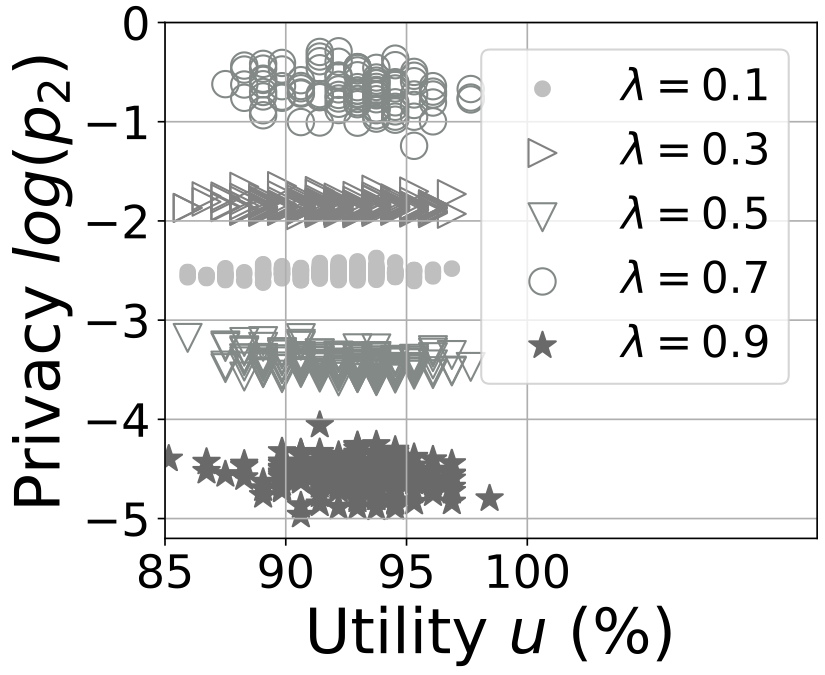}}
  \subfloat[PAN$_2$ on Har ($T_5$)]{
  \includegraphics[height=0.195\textwidth]{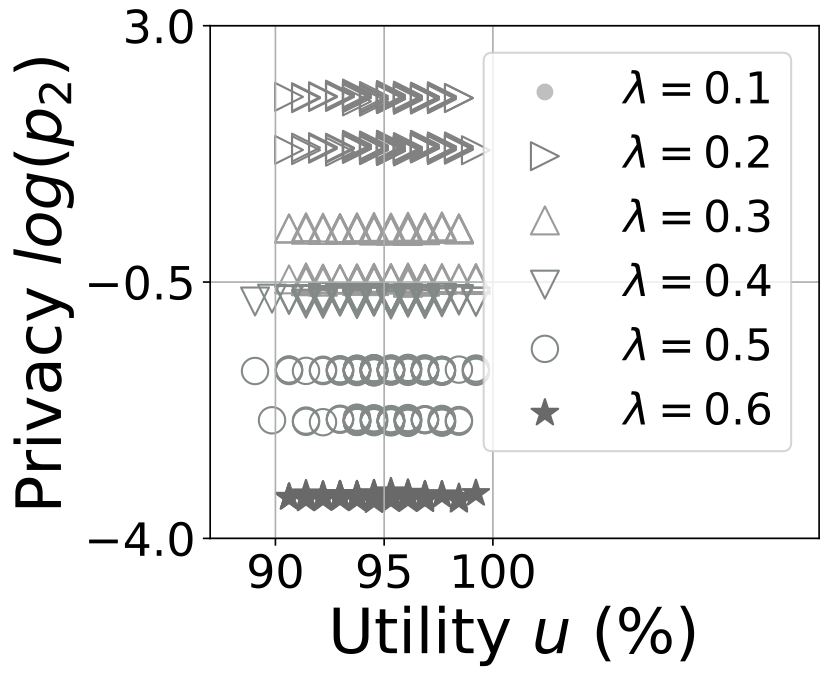}}
  \subfloat[PAN$_2$ on StateFarm ($T_6$)]{
  \includegraphics[height=0.195\textwidth]{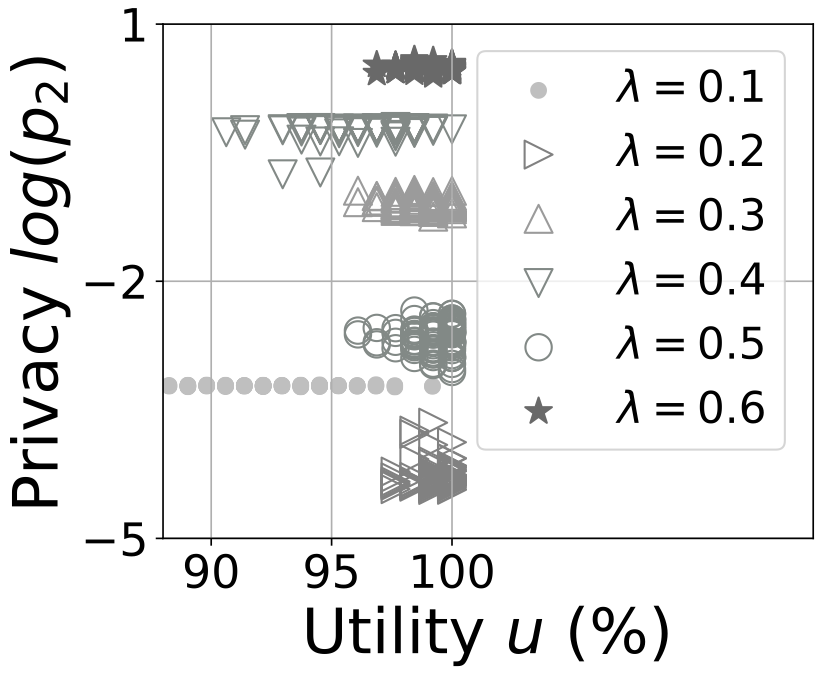}}
%\vspace{-4mm}
\caption{We can tune the utility-privacy tradeoff in both PAN$_1$ and PAN$_2$ by selecting different Lagrange multipliers $\lambda$ in the adversarial training phase (\equref{equ_csum}).}
\label{fig_lambda}
% \vspace{-4mm}
\end{figure}

\subsection{Impact of the Lagrangian Multipliers}
An important step in \systemnameposs training is to determine the Lagrangian multipliers $\lambda_1, \lambda_2, and \lambda_3$ in the adversarial training stage (see \equref{equ_csum}).
We verify that we are able to tune \systemnameposs utility-privacy tradeoff point through setting different $\lambda_1$, $\lambda_2$, and $\lambda_3$ in adversarila training phase (see \equref{equ_objective}), shown in Figure ~\ref{fig_lambda}.
Let $\lambda_3=0$ and $\lambda_2=1-\lambda_1$, we show evaluate the influence of Lagrangian multiplier on PAN$_1$ tradeoff performance over two typical applications (\eg $T_1$ and $T_4$), with five discrete choices of $\lambda_1$ $ \in \{0.1, 0.3, 0.5, 0.7, 0.9\}$.
As for PAN$_2$, empirically assuming $\lambda_3 = 0.2$ and $\lambda_2=1-0.2-\lambda_1$, we compare six discrete choices of $\lambda_1$ $ \in \{0.1, 0.2, 0.3, 0.4, 0.5, 0.6\}$.
% .
% \TODO{Check: [}
And we see that the optimal choice of Lagrange multiplier, among above optional space, is $\lambda_1=0.3$ for PAN$_1$ on digit classification ($T_1$: MNIST), is $\lambda_1=0.7$ for PAN$_1$ on non-speech sound recognition ($T_4$: Ubisound), is $\lambda_1=0.2$ for PAN$_2$ on human activity detection ($T_5$: Har), and $\lambda_1=0.6$ for PAN$_2$ on driver behavior classification ($T_6$: StateFarm).
%
% \TODO{]}

\vspace{1ex}\textbf{Summary.} 
The Lagrange multipliers $\lambda_1$, $\lambda_2$, and $\lambda_3$ bring flexibility to \systemname to satisfy different requirements of utility-privacy tradeoffs according to the relative importance between utility and privacy budgets across various tasks/applications.
And we note it is exhaustive to search the optimal $\lambda_1$, $\lambda_2$ and $\lambda_3$, since we can always search it from a finer-grained discrete space (\eg $\{0.0001, 0.0002, ..., 0.799\}$). An alternative in the future work is to leverage the automated search technique, \eg deep deterministic policy gradient algorithm, for efficient searching.
%
% For example, the feature generalization can be primarily enhanced by setting a larger balance coefficient $\lambda$.

\begin{table}[t]
\centering
\scriptsize
\caption{Resource cost of \systemnameposs Encoder for data encoding across five \lsc{Android applications} on Xiaomi Mi6 smartphone.}
%\vspace{-3mm}
\label{tb:cost}
\begin{tabular}{|c|c|c|c|}
\hline
% \multirow{2}{*}{Task} & \multicolumn{3}{l|}{s} \\ \cline{2-4} 
\multirow{2}{*}{\textbf{Applications}} & \multicolumn{3}{c|}{ \textbf{Encoder's Cost on Xiaomi Mi6 Smartphone}} \\ \cline{2-4} 
&\textbf{Latency ($ms$)} & \textbf{Storage ($KB$)} & \textbf{Energy ($mJ$)} \\ \hline
\textbf{Digit recognition ($T_1$: MNIST)} & 26 & 135 & 0.8 \\ \hline
\textbf{Image classification ($T_2$: CIFAR-10)} &  31 & 198 & 1.6 \\ \hline
\textbf{Image classification ($T_3$: ImageNet)} & 102 & 310 & 3.2 \\ \hline
\textbf{Acoustic event recognition ($T_4$: Ubisound)} &  42 & 213 & 1.8 \\ \hline
\textbf{Human activity prediction ($T_5$: Har)} & 27 & 269 & 0.9 \\ \hline
\end{tabular}
% \label{tb:cost}
% \vspace{-3mm}
\end{table}

\subsection{\lsc{Performance on Smartphone}}
We next evaluate \systemname on a commercial off-the-shelf smartphone with six Android applications. 

\subsubsection{Resource Cost of \systemnameposs Encoder on Smartphone}
\label{sec:cost}
% The high privacy and utility at the cost of .
This subsection evaluates the run-time execution cost (\eg latency, storage, and energy consumption) of the learned \systemnameposs Encoder for encoding different formats of data on the Xiaomi Mi6 smartphone. 
% \lin{either Xiaomi Mi6 or Redmi 6. Which one?}
\lsc{Specifically, we deploy the learned Encoder on the smartphone to interrupt and encode the incoming testing sample into features (\ie Encoder output). And then the Encoder output is fed into the corresponding Android Apps for task recognition and privacy validation. In this experiment, the task classifier is embedded in the corresponding Android App, and the privacy validation models (\ie private attribute classifier and privacy reconstructor) are executed on the cloud to attack the Encoder output collected by Android APP.} 
We summarize the on-device execution cost of \systemnameposs Encoder to encode five formats of data in Table \ref{tb:cost}.
%
% which are crucial budgets to deploy the \systemname on resource-limited ubiquitous platforms, especially when the Encoder is frequently invoked.
%
% shows the privacy and utility performance of feature generated by \systemnameposs Encoder, and the resource cost of \systemnameposs Encoder in terms of latency, energy cost, and storage occupation on .
%
In particular, we load the \systemnameposs Encoder (parameters and architecture files) in the smartphone cache to speedup processing, since it only occupies $\leq 310 KB$ storage. And the multiply-accumulate (MAC) operations of the Encoder network are run using smartphone CPU~\cite{bib:sicong2017:IMWUT}.
\systemnameposs Encoder occupies only $135 \sim 310 KB$ of memory, takes $26 \sim 102 ms$ of encoding latency, and incurs $0.8 \sim 3.2 mJ$ of energy cost for each encoding pass of raw data.

\vspace{1ex}\textbf{Summary.} \systemnameposs Encoder does not incur notable high resource cost. Therefore it is compact to deploy on the resource-constrained mobile platforms as a data preprocessing middleware. 
In particular, it takes low memory usage since the Encoder only contains convolutional layers, without storage-exhaustive fully-connected layers.
The execution delay is only several milliseconds~\cite{bib:liu2018:mobisys}.
And the energy cost is less than $\leq 3.2 mJ$, which is insignificant compared with Xiaomi Mi6's battery capacity, \ie $3350 mAh$.

\begin{table}[t]
\centering
\scriptsize
\caption{Performance on the driver behavior recognition Android App. Utility $u$ is the driver behavior recognition accuracy, specified privacy $p_1$ is the driver identity classification accuracy, and intuitive privacy $p_2$ is the data reconstruction error.}
%\vspace{-3mm}
\label{tb:case_study}
\begin{tabular}{|c|c|c|c|}
\hline
\textbf{Input to App} & Utility  $u$ (\%) & Specified privacy $p_1$ (\%) & Intuitive privacy $log(p_2)$ \\ \hline
Case A: Raw image & $98.2 $ & $93.6$ & $0$ \\ \hline
Case B: DNN features & $96.1 $ & $65.6$ & $0.035$ \\ \hline
Case C: \systemnameposs Encoder output & $99.1 $ & $23.1$ & $0.163 $ \\ \hline
\end{tabular}
%\vspace{-3mm}
\end{table}

\subsubsection{\lsc{Case studies on driver behavior recognition App}}

The user inputs data to an Android App to recognize driver behavior. Meanwhile, he wants to hide the private attributes (\eg the driver identity) and other agnostic private information (\eg the driver race and car model). Therefore, he leverages \systemname to encode the raw image into features and only deliver the Encoder output to the Android App.
We artificially play an example trace of the driver behavior during the study with $80$ driver images from $10$ drivers, selected from $4,424$ testing samples of StateFarm ($T_6$).
We consider 3 cases of the input data to the driver behavior recognition Android App: Case A: raw data, Case B: features generated by a standard DNN; and Case C: \systemnameposs Encoder output.
\tabref{tb:case_study} shows the evaluation results on the driver behavior recognition App for three cases.
In Case A with raw image input, the driver classification accuracy (utility) by App is $98.2\%$, while the adversary's accuracy of predicting the private attribute, \ie driver identity, is $93.6\%$. In Case B of DNN feature input, the utility of classifying driver behavior is $96.1\%$, and the private driver identity inference accuracy $p_1$ by the malicious attacker is $65.6\%$. It indicates that both the raw data and DNN feature cases reveal private attribute-correlated information. 
As for Case C with \systemnameposs Encoder output, it incurs an improvement ($0.9\%$) in driver behavior recognition accuracy, and reduce the private identity prediction accuracy $p_1$ by $70.5\%$.
Meanwhile, the intuitive reconstruction privacy $p_2$ in Case C is $0.163$, which is larger than that in Case B ($0$) and Case C ($0.035$).
The more significant reconstruction error implies the less private information leakage risk.

\vspace{1ex}\textbf{Summary.} This outcome demonstrates \systemnameposs Encoder improves utility with quantified privacy guarantees.

\begin{figure*}[t]
  \centering
  \subfloat[DNN]{
  \includegraphics[width=0.26\textwidth]{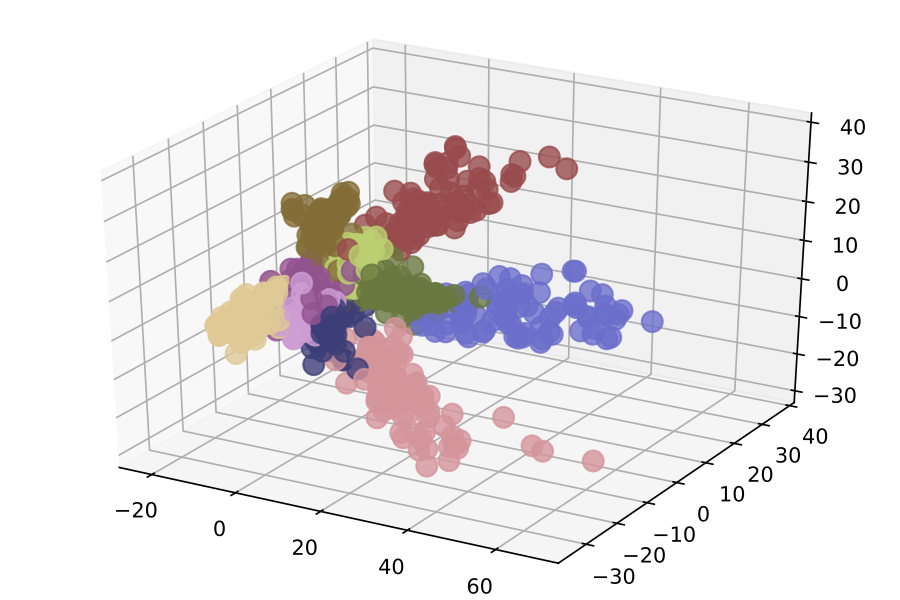}}
  \subfloat[Hybrid DNN]{
  \includegraphics[width=0.26\textwidth]{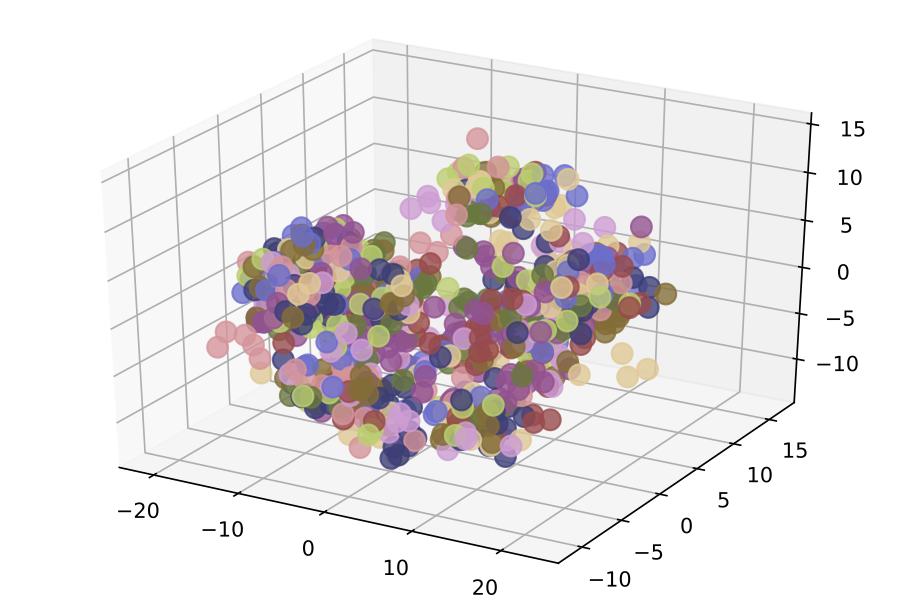}}
  \subfloat[PAN]{
  \includegraphics[width=0.26\textwidth]{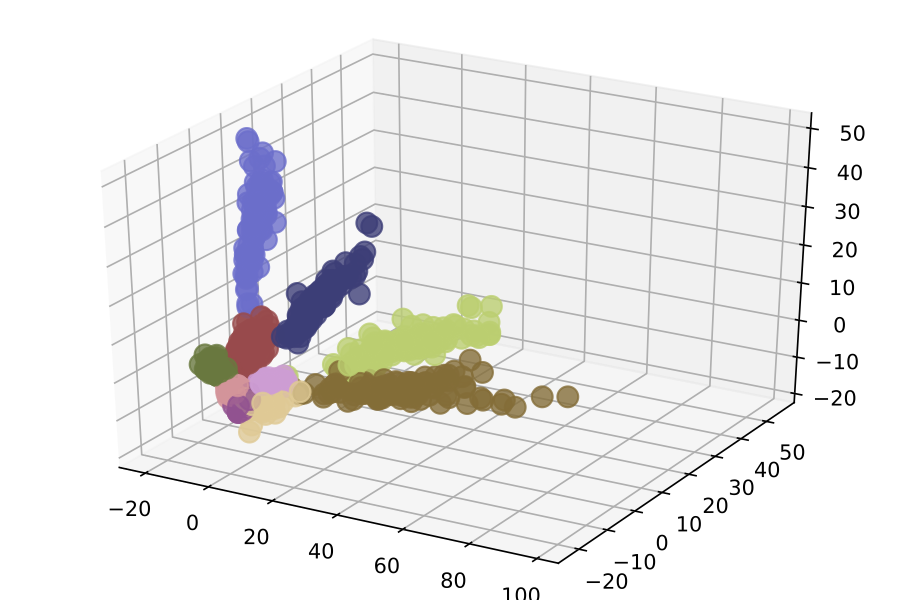}}
%\vspace{-3mm}
  \caption{Visualization of features learned by DNN, Hybrid method, and \systemnameposs Encoder on StateFarm ($T_6$) datasets. Different color in each figure standards for one task class.}
\label{fig_diff_feature}
\end{figure*}

\begin{figure*}[t]
  \centering
  \includegraphics[width=0.85\textwidth]{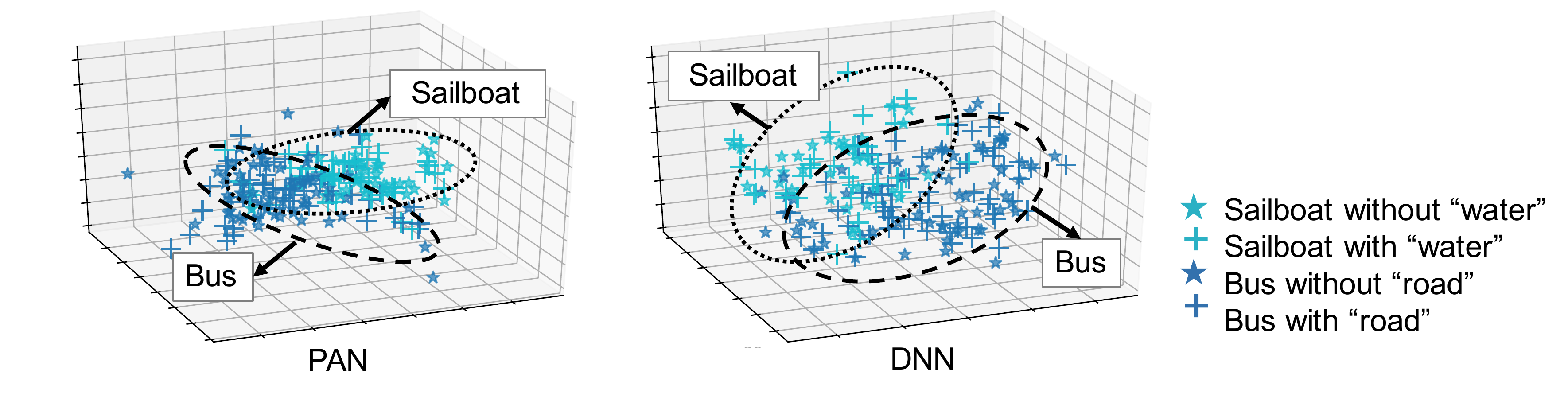}
  %\vspace{-3mm}
  \caption{Detail visualization of feature learning by PAN and DNN on two categories of images from ImageNet ($T_3$). \lsc{The target task is to classify the "sailboat" and "bus" image samples. The background "water" in "sailboat" images and the background "road" in "bus" images are redundant (private) to recognize the target "sailboat" and "bus"}.}
\label{fig_boat_feature}
\end{figure*}

\begin{figure*}[t]
  \centering
  \includegraphics[width=0.84\textwidth]{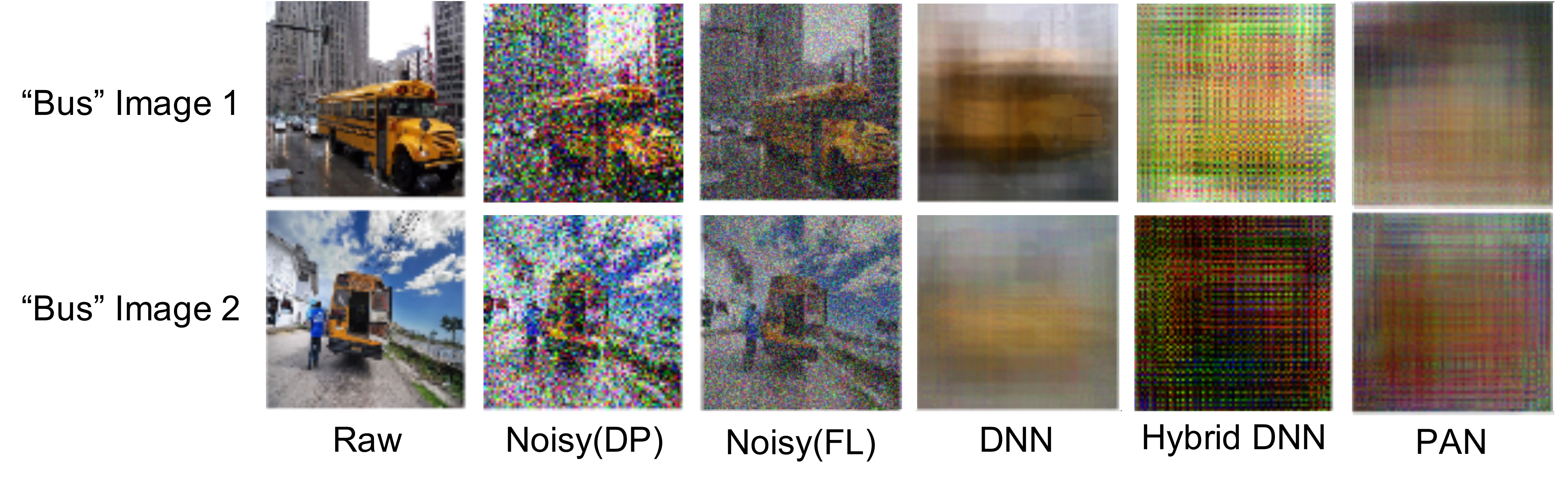}
  \vspace{-3mm}
  \caption{Visualization of reconstruction privacy. From left to right: raw "bus" images from ImageNet (Raw), images with \lsc{Laplace noise (DP), images with carefully-calibrated Gaussian noise (FL)}, and images reconstructed from DNN's features, Hybrid DNN's features, and PAN's Encoder output.}
\label{fig_diff_pixel}
%\vspace{-3mm}
\end{figure*}

\subsection{Visualization of \systemnameposs Encoder Output}
\label{subsec:visualize}
In this subsection, we further visualize the \systemnameposs Encoder output in terms of feature distribution and the reconstruction privacy \lsc{to seek insight into answering the following questions: what is the impact of \systemname on the learned features, how does \systemname disentangle the feature components relevant to privacy from those relevant to utility, and how well does \systemname preserve the reconstruction privacy of raw data}?

\subsubsection{Visualization of PAN's Encoder Output on Feature Space}
\label{sec:vis_utility}
Fig.~\ref{fig_diff_feature} and Fig.~\ref{fig_boat_feature} visualize how the feature manifold is derived by DNN, DNN(resized), and \systemname.
First, in Fig.~\ref{fig_diff_feature},  \systemnameposs Encoder output is highly separable as DNN method do on the feature space, which indicates its utility for task recognition.
\lsc{
While the manifold driven by the Hybrid baseline with PCA and noise addition processes on DNN features is blurry, this is why the resized features by Hybrid DNN method hurt utility.}
%f
Moreover, the feature distribution (manifold) formed by \systemname is the most constrictive one compared to that from the DNN and Hybrid baselines, \lsc{which leads to the improved utility.} 
% While the feature distribution from DNN(resized) is randomized.
%
Second, \lsc{\systemname to push the features away from redundant private information, for privacy, makes the manifold more constrictive, so that enhances the utility.} 
Specifically, to zoom in on two categories of images from ImageNet ($T_3$) for more details about how PAN and DNN form the feature manifold to achieve utility-privacy tradeoff, as shown in Fig.~\ref{fig_boat_feature}. 
\lsc{The target task is to classify the two categories, "sailboat" and "bus". The private background information in the "sailboat" raw image is "water", and the private information in the "bus" image is "road".}
We see \systemname pushes features towards the constrictive space dominated by the samples without redundant (private) information, \ie "sailboat without water" and "bus without road", \lsc{which guarantees privacy, avoids over-fitting, and improves utility as well.}
While the DNN method may capture the background (private) information "water" and "road" \lsc{and retain them in the feature manifold} to help the target task classification of "sailboat" and "bus", therefore hurts privacy.
%
% In this experiment, we take five typical mobile datasets, \ie MNIST (D1), CIFAR-10 (D2), ImageNet ($D_3$), Ubisound ($D_4$) and Har ($D_5$) as case studies, Fig.~\ref{fig_diff_feature} shows the distribution of the deep features learned by standard DNN and PAN.
%
% Furthermore, we find that the balance coefficients $\lambda$ in Eq.(4) has some influence on feature distribution direction and distribution contraction in the feature space.
%
% In summary, the highly separable features from \systemnameposs Encoder lead to a high utility on task inference. 
We defer the theoretical interpretation of this result to $\S$~ \ref{sec:discuss}.
% Second, the learning algorithm on PAN pushes features towards essential information and away from redundant background (sensitive) information (see more interpretations in ).

\subsubsection{Visualization of \systemnameposs Encoder output on Reconstruction Privacy}
\label{sec:vis_pivacy}
Fig.~\ref{fig_diff_pixel} visualizes the reconstruction privacy of \systemnameposs Encoder output, in comparison to the baseline approaches, using two "bus" image samples from ImageNet. 
%
% illustrates the pixel image of the raw data, the noisy data, the mimic data reconstructed from DNN's deep feature, and mimic data reconstructed from PAN's private features from two "bus" images from ImageNet datasets.
%
% In particular, the reconstructed data is the output of \systemnameposs Decoder based on the private features generated by \systemnameposs Encoder. 
%
We adopt the same architectures of encoder (\ie 12 conv layers, 5 pooling layers, and 1 batch-normalization layer) and privacy reconstructor (\ie the encoder turned upside down) to decode the features generated by DNN, Hybrid DNN, and \systemname for fair comparison.
%
% We leverage the reconstruction attacker (\ie decoder) to reconstruct images from the encoded features of DNN, DNN(resized) and PAN.
%
% We see the images reconstructed from \systemnameposs Encoder output are dramatically corrupted and hard to distinguish the exact information of sensitive information.
%
We see the images reconstructed from the DNN features convey \lsc{the target object "bus" information and the private background "road"} information, indicating a high risk of private background leakage.
Adding noise to the images in DP and FL or adding noise to the features in Hybrid DNN baselines obfuscate both \lsc{utility-related "bus" information} and \lsc{the privacy-correlated background "road" information}, compromising task detection accuracy (utility) at the cost of privacy. 
The \systemname, instead, only muddles the \lsc{utility-irrelevant private} information "road", making background information reconstruction impossible without compromising the utility. 

%% file: body/interpretation.tex
\section{Manifold based Interpretation}
\label{sec:discuss}
Our evaluation reported above shows that \systemname is able to train an Encoder that improves utility and accuracy at the same time. This section attempts to provide a theoretical interpretation of this surprising result.
%
% This section presents a theoretical interpretation to \systemnameposs design from the manifold learning perspective.
%
% Also, there are several limitations of this work that may need future exploration.
%
% \subsection{Manifold Based Interpretation}
%We resort to the manifold perspective of the deep model. 

We resort to the manifold perspective of the deep model. It is common in literature to assume that the high-dimensional raw data lies on a lower dimensional manifold~\cite{bib:chien2016:ICASSP}.
A DNN can also be viewed as a parametric manifold learner utilizing the nonlinear mapping of multi-layer architectures and connection weights.
%
%However, how to exactly characterize and control two types of descent gradients from discriminative error and generative error in back-propagation process to guide the manifold learning is still an open problem.
%
%In this paper, we argue that can we first learn the contractive manifold through standard generative or discriminative learning algorithm and then corrupt the manifold learning using an opposite generative learning objective? how can we force the manifold learning to improve the deep features' generalization ability and enhance the reconstruction difficulty based on the learned deep feature?
%
We decompose the input data into two orthogonal lower dimensional manifolds: $I =  I^{OD} +  I^{OD}_{\perp}$. Here, the component $I^{OD}$ is the manifold component that is both necessary and sufficient for task recognition (\eg driver behavior). 
Thus, ideally, we want our training algorithm to rely on this information for task recognition solely. 
Formally, for the utility discriminator (UD), this implies that $prob(y|I) = prob(y|I^{OD})$.
% ({\bf Anshu: Make sure the notation is consistent.})
%
And the other manifold component $I^{OD}_{\perp}$,
orthogonal to $I^{OD}$, may or may not contain information for the objective class, but it is dispensable for task detection. 
In practice, the real data does have redundant correlations. Thus $I^{OD}_{\perp}$ may be learned for task recognition, but unnecessary. 
However, revealing $I^{OD}_{\perp}$ is likely to contain some sensitive information (\eg driver identity information and background information) thus hurt the privacy.
If we assume that there does exist a sweet-spot tradeoff between utility and privacy, that we hope to find, then it must be the case that $I^{OD}$ is not sensitive. 

The features $F$ learned by standard discriminative learning to minimize the classification error based on information from $I$, will mostly likely overlap (non-zero projection) with both $I^{OD}$ and $I^{OD}_{\perp}$. And the overlap with $I^{OD}_{\perp}$ compromises the privacy (as evident from our experiments).
Meanwhile, the projection of manifold $I^{OD}$ on $I^{OD}_{\perp}$ is significant as it might capture other extra sensitive features, which will help task recognition accuracy.
Apart from privacy, the redundant correlation in $I^{OD}_{\perp}$ is also likely only be spurious in training data. Thus, merely minimizing classification loss can lead to over-fitting.

%Meanwhile, the projection of manifold $X^{OD}$ on $X^{OD}_{\perp}$ is significant as it might capture other extra sensitive features, which will help task recognition accuracy.
%

\begin{figure}[t]
\centering
\includegraphics[width=.9\textwidth]{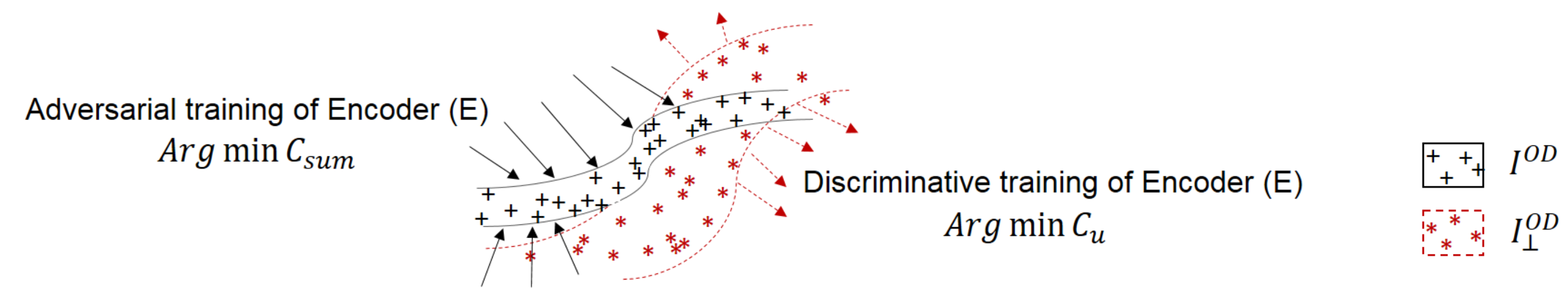}
%\vspace{-3mm}
\caption{A new manifold pushed by \systemname to form the feature extractor, \ie Encoder, for better utility and privacy.
The utility-specified discriminative learning objective (\equref{equ_cu}) push it to contain $I^{OD}$ and $I^{OD}_{\perp}$,
and the privacy-imposed adversarial training objective (\equref{equ_csum}) pushes it away from the sensitive component $I^{OD}_{\perp}$.
}
\label{fig_manifold}
%\vspace{-2mm}
\end{figure}

This is where we can skill two birds with one stone via an adversarial process. 
In \systemname, the Encoder $E(\cdot)$ is trained by the utility-specified discriminative learning objective (\equref{equ_cu}) and privacy-imposed adversarial learning objective (\equref{equ_csum}), to remove extra sensitive information in features $F'$ as shown in Fig.~\ref{fig_manifold}.
% \textbf{remove extra sensitive information in features}.
%
The transformed manifold formulated by Encoder $E(\cdot)$ is forced by discriminative learning objective (\equref{equ_cu}), just like the traditional approach, to contain information from both $I^{OD}$ as well as $I^{OD}_{\perp}$. 
However, the adversarial training objective (\equref{equ_csum}) will push features $F'$ away (or orthogonal) from $I^{OD}_{\perp}$. 
In this way, we get privacy as well, since $F'$ as a function of $I$ which has two manifolds, being orthogonal to $I^{OD}_{\perp}$ forces it to only depend on $I^{OD}$. 

Meanwhile, from a generalization perspective, 
in the training data, 
% these disturbing features do not harm or even improve object detection accuracy, because it has a regularizing effect.
% 
the spurious information from $I^{OD}_{\perp}$ that might over-fit the training data is iteratively removed by the adversarial training objective (\equref{equ_csum}), leading to enhanced generalization.
For example, as shown in Fig.~\ref{fig_boat_feature}, if we want to discriminate between "bus" and "sailboat", the background information "road" in the image can help in most cases but can also mislead when the test image contains a "sailboat" being transported on the "road". Therefore, by considering the background information, a standard DNN may not generalize well. 
In contrast, because the background may contain sensitive information and contribute to reconstruction error,  
\systemname is likely to train the Encoder to remove information about the background and as a result, improve the task accuracy .

The above interpretation highlights the possibility that utility and privacy are not completely competing objectives in practice. We believe that a rigorous formalism and thorough investigation of this phenomena is necessary to shed more insight and derive better designs.

% \rev{Will we improve or delete this  paragraph? It looks like a defective one.}
% Suppose there are a set of high-dimensional raw data $\mathcal{X}$ and a finite set of discrete label  $\mathcal{Y}$.
% %for supervised manifold learning.
% %
% We first optimally train the Encoder and Classifier's weights $\theta_e$ and $\theta_c$ to learn the regular intrinsic manifold $t(x)$ through a discriminative algorithm, which learns a conditional probability $P(y|x)$ of the class label $y$ given input data $x$, where $x$ and $y$ are samples in space $\mathcal{X}$ and $\mathcal{Y}$.
% %
% The class label $y$ is estimated by the Classifier's parametric function $C$ from the learned invariant manifold $t(x)$, denoted by $y=C(\theta_c, t(x))$.
% % 
% And then a generative algorithm focuses on how to generate mimic data from the learned feature, thus learns the joint distribution $P(t(x), y)$ and then compute the distribution of variable $t(x)$ and $y$ as marginal distribution $P(t(x))=\sum_{y}{P(t(x)|Y=y)}$ and 
% $P(Y)=\int_{t(x)}P(Y|X=t(x))$, respectively.
% %
% The conditional probability $P(t(x)|y)$ and $P(y|t(x))$ can be also computed by Bayes' rule from the joint probability $P(t(x),y)$ and the individual probability $P(y)$ or $P(t(x))$~\cite{bib:richard1991:NC}.
% %
% Next, the privacy and utility requirements-imposed adversarial training tunes above learning outcomes.

%% file: body/related.tex
\section{Related Work}
\label{sec:related}
Our work is inspired by and closely related to the following works. 

% \lin{First, how these works are related to RAN should be stated up front, not at the end!!! Second, you should highlight the DIFFERENCE/advantage of RAN, not just the similarity. Check my paper writing article again about how to discuss related work. Third, you did evaluate RAN against differential privacy (adding Laplace noise), right? You cite the result to bring out why RAN is different and better.}

% 素材

\textbf{Data Privacy Protection in Machine Learning based Services}:~~
Randomized noise addition~\cite{bib:he2017:ACC} and Differential privacy~\cite{bib:dwork2014:FTTCS,bib:abadi2016:abadi} techniques are widely used by service providers to remove personal identities in the released datasets.
They provide strong privacy guarantees but often lead to a significant reduction in utility (as shown in $\S$ \ref{subsec:exp_tradeoff}).
GAP~\cite{bib:huang:MDPI2017} learns a privatization scheme to sanitize the datasets for limiting the risk of inference attacks on personal private attributes. 
Erdogdu \etal design a privacy mapping scheme
for continuously released time-series of user data to protect the correlated private information in the dataset~\cite{bib:Erdogdu:AAAI2015}.
% \systemname is tailored to each piece of data, rather than the statistics of the datasets.
%
\lsc{
Federated learning~\cite{bib:truex:arXiv2018, bib:papernot:ICLR2018} techniques prevent inference over the sensitive data exchanged between distributed parties during training by noisy aggregation of multiple parties' resulting models.
%and ensuring utility by adding carefully-calibrated noise to the trained model.
}
However, all of the above techniques are tailored to datasets or the statistics of datasets, \lsc{which are unsuitable to our problem settings, \ie run-time data privacy protection in the online inference phase. Meanwhile, applying the statistic information to a new context-aware case is still an open problem.}

% thus inapplicable to our problem settings.
%
\lsc{
\systemname is a very different approach toward preserving the privacy at run-time. As the raw data is generated, it is intercepted by a trained Encoder and the encoded features are then fed into the untrusted service.}
% outperforms Noisy data (a differential privacy method) with better classification utility and competitive privacy ($\S$ 3.1), because \systemnameposs Encoder is end-to-end trained with collaborative utility-specified deep learning and privacy-imposed adversarial learning for a good trade-off between features' utility and privacy.

% 1. 为了保护隐私，我们传递特征，而不是原始数据
\textbf{Data Utility-Privacy Tradeoff using Adversarial Networks}:~~
Adversarial networks have been explored for data privacy protection, in which \rev{two or more players defend against others with conflicting utility/privacy goals}.
% \lin{``the privacy protector defends against the specific adversaries (attackers).'' does not make sense to me. Can you explain?}
%
Seong \etal \rev{introduce an adversarial game to learn the image obfuscation strategy, in which the user and recogniser (attacker) strive for antagonistic goals: dis-/enabling recognition~\cite{bib:oh:ICCV2017}.} 
% general game framework for adversarial image perturbations to confuse recognition systems.
% \lin{``recognition systems'' are attackers? How is their contribution? how did they find the image perturbations? Via adversarial training?}
%
Wu \etal ~\cite{bib:wu:ECCV2018} \rev{propose an adversarial framework to learn the degradation transformation (\eg anonymized video) of video inputs. This framework optimizes the tradeoff between task performance and privacy budgets.}
% \lin{Awkward English. Revise. Make the sentence simpler; break it into multiple shorter sentences if necessary.}
%
However, both practices only consider protecting privacy against \lsc{attackers that perform "vision" recognition on specific data format (\eg image), which is insufficient across diverse data modalities in ubiquitous computing. On the contrary, \systemname allow users to specify the utility and privacy quantification towards different data formats according to application requirements. And we have evaluated the \systemnameposs usability across image, audio, and motion data formats in $\S$~\ref{sec:experiment}.}
% \lin{Why is it insufficient? It implies your work is sufficient. How so? You need to briefly explain here.}
% 
OLYMPUS~\cite{bib:raval:pet2019} \rev{learns a data obfuscator (\ie AutoEncoder) to jointly minimize privacy and utility loss, where the privacy and utility requirements are modeled as adversarial networks.}
% \lin{Awkward English. Revise. Make the sentence simpler; break it into multiple shorter sentences if necessary.}
%

\lsc{Although the above works share the idea of adversarial learning with ours, they use the generative model to obfuscate the raw data in a homomorphic way. In contrast, \systemname uses the Encoder to learn a downsampling transformation, \eg, features, from the raw data, and send the features, rather than any forms of obfuscated/synthetic data, to service providers. A byproduct of this encoding is that PAN has more efficient data communication from the mobile to the service provider.}
% \lin{I added the last sentence. Please double check. Remove if it is not true.}

% 2. 处理特征比处理 data 更难！！ 因为特征处理是黑盒子，如果处理不好，可能损坏隐私，或者损坏可用性
\textbf{Deep Feature Learning for Utility or Privacy}:~~
%Learning features from raw data to enhance the service utility and suppress privacy leakage is challenging in the black-box of deep models.
% Xiang \etal introduce a linear method to disentangle the feature for facial identity and that for facial expression ~\cite{bib:xiang:TCSVT2018}.
Edwards \etal propose the adversarial learned representations that are both fair (independent of sensitive attributes) and discriminative for the prediction task~\cite{bib:edwards:arXiv2015}.
However, they target at fair decision \lsc{by quantifying the dependence between representation and sensitive variables thus provide no privacy guarantees}.
% make the decision be independent of sensitive variable (whether or not it contain private information).
%
Our work is closely related to \cite{bib:chen:CVPR2018} that employs a variational GAN to learn the representations that hide the personal identity and preserve the facial expression.
% disentangle the image representation from individual identity.
%
\lsc{However, it employs a generative model to minimize the reconstruction error for realistic image synthesis, which is vulnerable to agnostic privacy hacking by reverse engineering.
In contrast, we maximize the reconstruction error for intuitive privacy-preserving.}
%Meanwhile, 
%
Also, discriminative and generative models are widely studied for latent feature learning, improving task inference but facilitating data reconstruction~\cite{bib:radford2015:arxiv, bib:zhong2016:FDS}. \lsc{They would make intuitive privacy protection even harder.}
%
% And some components used in existing generative models, such as sensitivity penalty in contractive autoencoder~\cite{bib:rifai2011:ICML}, data probability distribution in generative adversarial network~\cite{bib:goodfellow2014:advances} and KL divergence in variational autoencoder~\cite{bib:doersch2016:arxiv}, can be further integrated into \systemnameposs framework to define and enhance application-based privacy.
%
% \systemname 
% is the first to concurrently achieve the service-specified utility, suppress the user-specified privacy and the agnostic privacy leakage against any considerable privacy attackers. 
Osia \etal~\cite{bib:ossia17:arxiv} employ a combination of dimensionality reduction, noise addition, and Siamese fine-tuning to preserve privacy.
Importantly both its dimensionality reduction and Siamese fine-tuning are based on discriminative training. Specifically its Siamese fine-tuning seeks to reduce the intra-class variation in features amongst training samples for the intended classification service.
While the authors show these methods improve privacy, there is no systematic way to make tradeoffs between privacy and utility. In contrast, \systemname presents a rigorous mechanism to discover good tradeoffs via a combination of discriminative, generative, and adversarial training.
Mohammad \etal~\cite{bib:malekzadeh:IoTDI2018} present a privacy-preserving transformation called Replacement AutoEncoder (RAE). Like the Encoder in \systemname, RAE also intends to eliminate sensitive information from the features while keeping the task-relevant information.
Importantly it assumes that features/data relevant to an intended task do not overlap with those revealing sensitive information. 
As a result, it transforms the data by simply replacing the latter with features/data that are irrelevant to the intended task and do not reveal sensitive information.
With that assumption, RAE eschews the hard problem of making a good tradeoff between utility and privacy and is also solely based on discriminative training.
Furthermore, RAE does not reduce the amount of data that have to be sent to the service provider and would use significantly higher resources in transforming the data in which RAE will need to detect sensitive features/data, replace them, and then reconstruct the (modified) raw data. In contrast, \systemname only needs to run the dimensionality-reducing Encoder on the raw data and send the features to the service provider, although \systemname may require significantly more computational resources in training the Encoder, which is done off-line, in the cloud. 

%% file: body/conclusion.tex
\section{Concluding Remarks}
\label{sec:conclude}
This paper addresses the privacy concern when mobile users send their data to an untrusted service provider for classification services.
We present \systemname, an adversarial framework to automatically generate deep features from the raw data with quantified guarantees in privacy and utility.
%
%\systemname is the first to learn the feature extractor, i.e., Encoder (E), to explicitly learn features for maximized task inference accuracy, minimized privacy recognition accuracy, and maximized privacy reconstruction error via an adversarial training process.
%
We report a prototype of \systemname on Android platforms and cloud servers.
Evaluation using \lsc{Android applications and benchmark datasets} show that \systemnameposs Encoder output attains a notably better privacy-utility tradeoff than known methods. To our surprise, it achieves even better utility than standard DNNs that are completely ignorant of privacy.
We surmise that this surprising result can be understood from the perspective of manifold.
%
% In particular, \systemnameposs Encoder (E) defend against multiple adversaries.
% consists an Encoder (E) for feature extracting, a privacy reconstructor (PR) for data reconstruction error (privacy) strive for
% quantification from Encoder output and a Classifier for accuracy (utility) discrimination.
%
% The proposed training algorithm upon \systemnameposs contains three phase: discriminative learning function on Encoder and Classifier to boost their discriminative abilities, a generative stage on Decoder to improve its data generative capacity which stand in the position of Encoder's adversary, and an adversarial stage on Encoder, Classifier and Decoder to achieve our design objectives.
%
% All of these three parts are trained by service provider and only the trained Encoder is deployed on mobile platform to be a data privacy-preserving tool which is totally under mobile users' control.
%

%
% In the future, we plan to investigate finer-grained manifold learning techniques on \systemname for feature generalization and privacy improvements.

We also see three directions that the work reported in this paper can be extended.
% First, \systemname .
First, the \systemname framework can accommodate other choices of \lsc{context-aware utility, such as the sequence prediction, and privacy quantification, such as the information theory-based privacy, according to app requirements. 
%
% For example, the theoreti
%
Second, it can also integrate multiple utility discriminators and privacy attackers to train the Encoder, given the appropriate datasets accompanied by utility and privacy labels.}
%
% And we can also measure the privacy by the hidden failure, \ie the ratio between the background patterns that were discovered based on \systemnameposs Encoder output, and the sensitive patterns founded from the raw data, in an object recognition task.
%
Third, our experience shows that the training of multiple adversarial models in \systemname must be carefully synchronized to avoid model degradation caused by difference in their objectives and convergence speeds.
Therefore, more heuristics and insights for guaranteeing and accelerating training convergence are much needed.

%% file: PAN.bbl

\begin{thebibliography}{43}

%%% the Web address.

\ifx \showCODEN    \undefined \def \showCODEN     #1{\unskip}     \fi
\ifx \showDOI      \undefined \def \showDOI       #1{#1}\fi
\ifx \showISBNx    \undefined \def \showISBNx     #1{\unskip}     \fi
\ifx \showISBNxiii \undefined \def \showISBNxiii  #1{\unskip}     \fi
\ifx \showISSN     \undefined \def \showISSN      #1{\unskip}     \fi
\ifx \showLCCN     \undefined \def \showLCCN      #1{\unskip}     \fi
\ifx \shownote     \undefined \def \shownote      #1{#1}          \fi
\ifx \showarticletitle \undefined \def \showarticletitle #1{#1}   \fi
\ifx \showURL      \undefined \def \showURL       {\relax}        \fi
% invisible to TeX
\providecommand\bibfield[2]{#2}
\providecommand\bibinfo[2]{#2}
\providecommand\natexlab[1]{#1}
\providecommand\showeprint[2][]{arXiv:#2}

\bibitem[\protect\citeauthoryear{Abadi, Chu, Goodfellow, McMahan, Mironov,
  Talwar, and Zhang}{Abadi et~al\mbox{.}}{2016}]%
        {bib:abadi2016:abadi}
\bibfield{author}{\bibinfo{person}{Martin Abadi}, \bibinfo{person}{Andy Chu},
  \bibinfo{person}{Ian Goodfellow}, \bibinfo{person}{H~Brendan McMahan},
  \bibinfo{person}{Ilya Mironov}, \bibinfo{person}{Kunal Talwar}, {and}
  \bibinfo{person}{Li Zhang}.} \bibinfo{year}{2016}\natexlab{}.
\newblock \showarticletitle{Deep learning with differential privacy}.
\newblock In \bibinfo{booktitle}{\emph{Proceedings of SIGSAC}}.
  \bibinfo{pages}{308--318}.
\newblock


\bibitem[\protect\citeauthoryear{Bhatia, Breaux, Friedberg, Hibshi, and
  Smullen}{Bhatia et~al\mbox{.}}{2016}]%
        {bhatia2016privacy}
\bibfield{author}{\bibinfo{person}{Jaspreet Bhatia}, \bibinfo{person}{Travis~D
  Breaux}, \bibinfo{person}{Liora Friedberg}, \bibinfo{person}{Hanan Hibshi},
  {and} \bibinfo{person}{Daniel Smullen}.} \bibinfo{year}{2016}\natexlab{}.
\newblock \showarticletitle{Privacy risk in cybersecurity data sharing}. In
  \bibinfo{booktitle}{\emph{Proceedings of ACM Workshop on ISCS}}. ACM,
  \bibinfo{pages}{57--64}.
\newblock


\bibitem[\protect\citeauthoryear{Chen, Konrad, and Ishwar}{Chen
  et~al\mbox{.}}{2018}]%
        {bib:chen:CVPR2018}
\bibfield{author}{\bibinfo{person}{Jiawei Chen}, \bibinfo{person}{Janusz
  Konrad}, {and} \bibinfo{person}{Prakash Ishwar}.}
  \bibinfo{year}{2018}\natexlab{}.
\newblock \showarticletitle{Vgan-based image representation learning for
  privacy-preserving facial expression recognition}. In
  \bibinfo{booktitle}{\emph{Proceedings of CVPR Workshops}}.
  \bibinfo{pages}{1570--1579}.
\newblock


\bibitem[\protect\citeauthoryear{Chien and Chen}{Chien and Chen}{2016}]%
        {bib:chien2016:ICASSP}
\bibfield{author}{\bibinfo{person}{Jen-Tzung Chien} {and}
  \bibinfo{person}{Ching-Huai Chen}.} \bibinfo{year}{2016}\natexlab{}.
\newblock \showarticletitle{Deep discriminative manifold learning}.
\newblock In \bibinfo{booktitle}{\emph{Proceeding of ICASSP}}.
  \bibinfo{pages}{2672--2676}.
\newblock


\bibitem[\protect\citeauthoryear{Collette}{Collette}{2018}]%
        {lib:h5py}
\bibfield{author}{\bibinfo{person}{Andrew Collette}.}
  \bibinfo{year}{2018}\natexlab{}.
\newblock \bibinfo{title}{HDF5 for Python}.
\newblock \bibinfo{howpublished}{\url{http://www.h5py.org/}}.
\newblock


\bibitem[\protect\citeauthoryear{Deng, Dong, Socher, Li, Li, and Fei-Fei}{Deng
  et~al\mbox{.}}{2009}]%
        {data:imagenet}
\bibfield{author}{\bibinfo{person}{Jia Deng}, \bibinfo{person}{Wei Dong},
  \bibinfo{person}{Richard Socher}, \bibinfo{person}{Li-Jia Li},
  \bibinfo{person}{Kai Li}, {and} \bibinfo{person}{Li Fei-Fei}.}
  \bibinfo{year}{2009}\natexlab{}.
\newblock \showarticletitle{Imagenet: A large-scale hierarchical image
  database}.
\newblock In \bibinfo{booktitle}{\emph{Proceedings of CVPR}}.
\newblock


\bibitem[\protect\citeauthoryear{Dwork, Naor, Pitassi, and Rothblum}{Dwork
  et~al\mbox{.}}{2010}]%
        {bib:dwork2010:STC}
\bibfield{author}{\bibinfo{person}{Cynthia Dwork}, \bibinfo{person}{Moni Naor},
  \bibinfo{person}{Toniann Pitassi}, {and} \bibinfo{person}{Guy~N Rothblum}.}
  \bibinfo{year}{2010}\natexlab{}.
\newblock \showarticletitle{Differential privacy under continual observation}.
  In \bibinfo{booktitle}{\emph{Proceedings of STC}}. ACM,
  \bibinfo{pages}{715--724}.
\newblock


\bibitem[\protect\citeauthoryear{Dwork, Roth, et~al\mbox{.}}{Dwork
  et~al\mbox{.}}{2014}]%
        {bib:dwork2014:FTTCS}
\bibfield{author}{\bibinfo{person}{Cynthia Dwork}, \bibinfo{person}{Aaron
  Roth}, {et~al\mbox{.}}} \bibinfo{year}{2014}\natexlab{}.
\newblock \showarticletitle{The algorithmic foundations of differential
  privacy}.
\newblock \bibinfo{journal}{\emph{Journal of Foundations and Trends in
  Theoretical Computer Science}} (\bibinfo{year}{2014}),
  \bibinfo{pages}{211--407}.
\newblock


\bibitem[\protect\citeauthoryear{Dwork, Smith, Steinke, and Ullman}{Dwork
  et~al\mbox{.}}{2017}]%
        {bib:dwork2017:ARSA}
\bibfield{author}{\bibinfo{person}{Cynthia Dwork}, \bibinfo{person}{Adam
  Smith}, \bibinfo{person}{Thomas Steinke}, {and} \bibinfo{person}{Jonathan
  Ullman}.} \bibinfo{year}{2017}\natexlab{}.
\newblock \showarticletitle{Exposed! a survey of attacks on private data}.
\newblock \bibinfo{journal}{\emph{Annual Review of Statistics and Its
  Application}}  \bibinfo{volume}{4} (\bibinfo{year}{2017}),
  \bibinfo{pages}{61--84}.
\newblock


\bibitem[\protect\citeauthoryear{Edwards and Storkey}{Edwards and
  Storkey}{2015}]%
        {bib:edwards:arXiv2015}
\bibfield{author}{\bibinfo{person}{Harrison Edwards} {and}
  \bibinfo{person}{Amos Storkey}.} \bibinfo{year}{2015}\natexlab{}.
\newblock \showarticletitle{Censoring representations with an adversary}.
\newblock \bibinfo{journal}{\emph{arXiv preprint arXiv:1511.05897}}
  (\bibinfo{year}{2015}).
\newblock


\bibitem[\protect\citeauthoryear{Erdogdu, Fawaz, and Montanari}{Erdogdu
  et~al\mbox{.}}{2015}]%
        {bib:Erdogdu:AAAI2015}
\bibfield{author}{\bibinfo{person}{Murat~A Erdogdu}, \bibinfo{person}{Nadia
  Fawaz}, {and} \bibinfo{person}{Andrea Montanari}.}
  \bibinfo{year}{2015}\natexlab{}.
\newblock \showarticletitle{Privacy-utility trade-off for time-series with
  application to smart-meter data}. In \bibinfo{booktitle}{\emph{Proceedings of
  Workshops at AAAI}}.
\newblock


\bibitem[\protect\citeauthoryear{Giusti, Ciresan, Masci, Gambardella, and
  Schmidhuber}{Giusti et~al\mbox{.}}{2013}]%
        {bib:giusti2013:ICIP}
\bibfield{author}{\bibinfo{person}{Alessandro Giusti}, \bibinfo{person}{Dan~C
  Ciresan}, \bibinfo{person}{Jonathan Masci}, \bibinfo{person}{Luca~M
  Gambardella}, {and} \bibinfo{person}{Jurgen Schmidhuber}.}
  \bibinfo{year}{2013}\natexlab{}.
\newblock \showarticletitle{Fast image scanning with deep max-pooling
  convolutional neural networks}.
\newblock In \bibinfo{booktitle}{\emph{Proceedings of ICIP}}.
  \bibinfo{pages}{4034--4038}.
\newblock


\bibitem[\protect\citeauthoryear{Goodfellow, Pouget-Abadie, Mirza, Xu,
  Warde-Farley, Ozair, Courville, and Bengio}{Goodfellow et~al\mbox{.}}{2014}]%
        {bib:goodfellow2014:advances}
\bibfield{author}{\bibinfo{person}{Ian Goodfellow}, \bibinfo{person}{Jean
  Pouget-Abadie}, \bibinfo{person}{Mehdi Mirza}, \bibinfo{person}{Bing Xu},
  \bibinfo{person}{David Warde-Farley}, \bibinfo{person}{Sherjil Ozair},
  \bibinfo{person}{Aaron Courville}, {and} \bibinfo{person}{Yoshua Bengio}.}
  \bibinfo{year}{2014}\natexlab{}.
\newblock \showarticletitle{Generative adversarial nets}.
\newblock In \bibinfo{booktitle}{\emph{Advances in Neural Information
  Processing Systems}}. \bibinfo{pages}{2672--2680}.
\newblock


\bibitem[\protect\citeauthoryear{Google}{Google}{2018a}]%
        {bib:LruCache}
\bibfield{author}{\bibinfo{person}{Google}.} \bibinfo{year}{2018}\natexlab{a}.
\newblock \bibinfo{title}{android.util.LruCache}.
\newblock
  \bibinfo{howpublished}{\url{https://developer.android.com/reference/android/util/LruCache.html}}.
\newblock


\bibitem[\protect\citeauthoryear{Google}{Google}{2018b}]%
        {lib:tf:mobile}
\bibfield{author}{\bibinfo{person}{Google}.} \bibinfo{year}{2018}\natexlab{b}.
\newblock \bibinfo{title}{TensorFlow Mobile}.
\newblock \bibinfo{howpublished}{\url{https://www.tensorflow.org/mobile/}}.
\newblock


\bibitem[\protect\citeauthoryear{GoogleCloud}{GoogleCloud}{2018}]%
        {url:googlecloud}
\bibfield{author}{\bibinfo{person}{GoogleCloud}.}
  \bibinfo{year}{2018}\natexlab{}.
\newblock \bibinfo{title}{Data Preparation}.
\newblock
  \bibinfo{howpublished}{\url{https://cloud.google.com/ml-engine/docs/tensorflow/data-prep}}.
\newblock


\bibitem[\protect\citeauthoryear{GoogleNow}{GoogleNow}{2018}]%
        {url:googlenow}
\bibfield{author}{\bibinfo{person}{GoogleNow}.}
  \bibinfo{year}{2018}\natexlab{}.
\newblock \bibinfo{title}{Google Now Launcher}.
\newblock
  \bibinfo{howpublished}{\url{https://en.wikipedia.org/wiki/Google_Now}}.
\newblock


\bibitem[\protect\citeauthoryear{He and Cai}{He and Cai}{2017}]%
        {bib:he2017:ACC}
\bibfield{author}{\bibinfo{person}{Jianping He} {and} \bibinfo{person}{Lin
  Cai}.} \bibinfo{year}{2017}\natexlab{}.
\newblock \showarticletitle{Differential private noise adding mechanism: Basic
  conditions and its application}. In \bibinfo{booktitle}{\emph{American
  Control Conference (ACC), 2017}}. IEEE, \bibinfo{pages}{1673--1678}.
\newblock


\bibitem[\protect\citeauthoryear{Huang, Kairouz, Chen, Sankar, and
  Rajagopal}{Huang et~al\mbox{.}}{2017}]%
        {bib:huang:MDPI2017}
\bibfield{author}{\bibinfo{person}{Chong Huang}, \bibinfo{person}{Peter
  Kairouz}, \bibinfo{person}{Xiao Chen}, \bibinfo{person}{Lalitha Sankar},
  {and} \bibinfo{person}{Ram Rajagopal}.} \bibinfo{year}{2017}\natexlab{}.
\newblock \showarticletitle{Context-aware generative adversarial privacy}.
\newblock \bibinfo{journal}{\emph{Entropy}} (\bibinfo{year}{2017}).
\newblock


\bibitem[\protect\citeauthoryear{Ioffe and Szegedy}{Ioffe and Szegedy}{2015}]%
        {bib:ioffe2015:arxiv}
\bibfield{author}{\bibinfo{person}{Sergey Ioffe} {and}
  \bibinfo{person}{Christian Szegedy}.} \bibinfo{year}{2015}\natexlab{}.
\newblock \showarticletitle{Batch normalization: Accelerating deep network
  training by reducing internal covariate shift}.
\newblock \bibinfo{journal}{\emph{arXiv preprint arXiv:1502.03167}}
  (\bibinfo{year}{2015}).
\newblock


\bibitem[\protect\citeauthoryear{Kaggle}{Kaggle}{2019}]%
        {data:statefarm}
\bibfield{author}{\bibinfo{person}{Kaggle}.} \bibinfo{year}{2019}\natexlab{}.
\newblock \bibinfo{title}{State Farm Distracted Driver Detection}.
\newblock
  \bibinfo{howpublished}{\url{https://www.kaggle.com/c/state-farm-distracted-driver-detection}}.
\newblock


\bibitem[\protect\citeauthoryear{Kingma and Ba}{Kingma and Ba}{2014}]%
        {bib:kingma2014:arxiv}
\bibfield{author}{\bibinfo{person}{Diederik~P Kingma} {and}
  \bibinfo{person}{Jimmy Ba}.} \bibinfo{year}{2014}\natexlab{}.
\newblock \showarticletitle{Adam: A method for stochastic optimization}.
\newblock \bibinfo{journal}{\emph{arXiv preprint arXiv:1412.6980}}
  (\bibinfo{year}{2014}).
\newblock


\bibitem[\protect\citeauthoryear{Krizhevsky, Vinod, and Geoffrey}{Krizhevsky
  et~al\mbox{.}}{2014}]%
        {data:cifar}
\bibfield{author}{\bibinfo{person}{Alex Krizhevsky}, \bibinfo{person}{Nair
  Vinod}, {and} \bibinfo{person}{Hinton Geoffrey}.}
  \bibinfo{year}{2014}\natexlab{}.
\newblock \bibinfo{title}{The CIFAR-10 dataset}.
\newblock \bibinfo{howpublished}{\url{https://goo.gl/hXmru5}}.
\newblock


\bibitem[\protect\citeauthoryear{Kruse, Borgelt, Klawonn, Moewes, Steinbrecher,
  and Held}{Kruse et~al\mbox{.}}{2013}]%
        {bib:kruse2013:CI}
\bibfield{author}{\bibinfo{person}{Rudolf Kruse}, \bibinfo{person}{Christian
  Borgelt}, \bibinfo{person}{Frank Klawonn}, \bibinfo{person}{Christian
  Moewes}, \bibinfo{person}{Matthias Steinbrecher}, {and}
  \bibinfo{person}{Pascal Held}.} \bibinfo{year}{2013}\natexlab{}.
\newblock \showarticletitle{Multi-layer perceptrons}.
\newblock \bibinfo{publisher}{Springer}, \bibinfo{pages}{47--81}.
\newblock


\bibitem[\protect\citeauthoryear{LeCun}{LeCun}{1998}]%
        {data:mnist1998:LeCun}
\bibfield{author}{\bibinfo{person}{Yann LeCun}.}
  \bibinfo{year}{1998}\natexlab{}.
\newblock \bibinfo{title}{The MNIST database of handwritten digits}.
\newblock \bibinfo{howpublished}{\url{https://goo.gl/t6gTEy}}.
\newblock


\bibitem[\protect\citeauthoryear{Li, Zhang, Chen, and Smola}{Li
  et~al\mbox{.}}{2014}]%
        {bib:li2014:sigkdd}
\bibfield{author}{\bibinfo{person}{Mu Li}, \bibinfo{person}{Tong Zhang},
  \bibinfo{person}{Yuqiang Chen}, {and} \bibinfo{person}{Alexander~J Smola}.}
  \bibinfo{year}{2014}\natexlab{}.
\newblock \showarticletitle{Efficient mini-batch training for stochastic
  optimization}. In \bibinfo{booktitle}{\emph{Proceedings of SIGKDD}}. ACM,
  \bibinfo{pages}{661--670}.
\newblock


\bibitem[\protect\citeauthoryear{Liu, Lin, Zhou, Nan, Liu, and Du}{Liu
  et~al\mbox{.}}{2018}]%
        {bib:liu2018:mobisys}
\bibfield{author}{\bibinfo{person}{Sicong Liu}, \bibinfo{person}{Yingyan Lin},
  \bibinfo{person}{Zimu Zhou}, \bibinfo{person}{Kaiming Nan},
  \bibinfo{person}{Hui Liu}, {and} \bibinfo{person}{Junzhao Du}.}
  \bibinfo{year}{2018}\natexlab{}.
\newblock \showarticletitle{On-demand deep model compression for mobile
  devices: a usage-driven model selection framework}.
\newblock In \bibinfo{booktitle}{\emph{Proceedings of ACM MobiSys}}.
\newblock


\bibitem[\protect\citeauthoryear{Mahendran and Vedaldi}{Mahendran and
  Vedaldi}{2015}]%
        {bib:mahendran2015:CVPR}
\bibfield{author}{\bibinfo{person}{Aravindh Mahendran} {and}
  \bibinfo{person}{Andrea Vedaldi}.} \bibinfo{year}{2015}\natexlab{}.
\newblock \bibinfo{title}{Understanding deep image representations by inverting
  them}.
\newblock , \bibinfo{numpages}{5188--5196}~pages.
\newblock


\bibitem[\protect\citeauthoryear{Malekzadeh, Clegg, and Haddadi}{Malekzadeh
  et~al\mbox{.}}{2018}]%
        {bib:malekzadeh:IoTDI2018}
\bibfield{author}{\bibinfo{person}{Mohammad Malekzadeh},
  \bibinfo{person}{Richard~G Clegg}, {and} \bibinfo{person}{Hamed Haddadi}.}
  \bibinfo{year}{2018}\natexlab{}.
\newblock \showarticletitle{Replacement autoencoder: A privacy-preserving
  algorithm for sensory data analysis}. In
  \bibinfo{booktitle}{\emph{Proceedings of IEEE IoTDI}}.
  \bibinfo{pages}{165--176}.
\newblock


\bibitem[\protect\citeauthoryear{Mendes and Vilela}{Mendes and Vilela}{2017}]%
        {bib:mendes2017:access}
\bibfield{author}{\bibinfo{person}{Ricardo Mendes} {and}
  \bibinfo{person}{Jo{\~a}o~P Vilela}.} \bibinfo{year}{2017}\natexlab{}.
\newblock \showarticletitle{Privacy-preserving data mining: methods, metrics,
  and applications}.
\newblock \bibinfo{journal}{\emph{IEEE Access}}  \bibinfo{volume}{5}
  (\bibinfo{year}{2017}), \bibinfo{pages}{10562--10582}.
\newblock


\bibitem[\protect\citeauthoryear{Milletari, Navab, and Ahmadi}{Milletari
  et~al\mbox{.}}{2016}]%
        {bib:milletari2016:3DV}
\bibfield{author}{\bibinfo{person}{Fausto Milletari}, \bibinfo{person}{Nassir
  Navab}, {and} \bibinfo{person}{Seyed-Ahmad Ahmadi}.}
  \bibinfo{year}{2016}\natexlab{}.
\newblock \showarticletitle{V-net: Fully convolutional neural networks for
  volumetric medical image segmentation}.
\newblock In \bibinfo{booktitle}{\emph{Proceedings of 3DV}}.
  \bibinfo{pages}{565--571}.
\newblock


\bibitem[\protect\citeauthoryear{Oh, Fritz, and Schiele}{Oh
  et~al\mbox{.}}{2017}]%
        {bib:oh:ICCV2017}
\bibfield{author}{\bibinfo{person}{Seong~Joon Oh}, \bibinfo{person}{Mario
  Fritz}, {and} \bibinfo{person}{Bernt Schiele}.}
  \bibinfo{year}{2017}\natexlab{}.
\newblock \showarticletitle{Adversarial image perturbation for privacy
  protection a game theory perspective}. In
  \bibinfo{booktitle}{\emph{Proceedings of ICCV}}. \bibinfo{pages}{1491--1500}.
\newblock


\bibitem[\protect\citeauthoryear{Ossia, Shamsabadi, Taheri, Rabiee, Lane, and
  Haddadi}{Ossia et~al\mbox{.}}{2017}]%
        {bib:ossia17:arxiv}
\bibfield{author}{\bibinfo{person}{Seyed~Ali Ossia},
  \bibinfo{person}{Ali~Shahin Shamsabadi}, \bibinfo{person}{Ali Taheri},
  \bibinfo{person}{Hamid~R Rabiee}, \bibinfo{person}{Nic Lane}, {and}
  \bibinfo{person}{Hamed Haddadi}.} \bibinfo{year}{2017}\natexlab{}.
\newblock \showarticletitle{A hybrid deep learning architecture for
  privacy-preserving mobile analytics}.
\newblock \bibinfo{journal}{\emph{arXiv preprint arXiv:1703.02952}}
  (\bibinfo{year}{2017}).
\newblock


\bibitem[\protect\citeauthoryear{Papernot, Song, Mironov, Raghunathan, Talwar,
  and Erlingsson}{Papernot et~al\mbox{.}}{2018}]%
        {bib:papernot:ICLR2018}
\bibfield{author}{\bibinfo{person}{Nicolas Papernot}, \bibinfo{person}{Shuang
  Song}, \bibinfo{person}{Ilya Mironov}, \bibinfo{person}{Ananth Raghunathan},
  \bibinfo{person}{Kunal Talwar}, {and} \bibinfo{person}{{\'U}lfar
  Erlingsson}.} \bibinfo{year}{2018}\natexlab{}.
\newblock \showarticletitle{Scalable private learning with pate}.
\newblock \bibinfo{journal}{\emph{Proceddings of ICLR}} (\bibinfo{year}{2018}).
\newblock


\bibitem[\protect\citeauthoryear{Radford, Metz, and Chintala}{Radford
  et~al\mbox{.}}{2015}]%
        {bib:radford2015:arxiv}
\bibfield{author}{\bibinfo{person}{Alec Radford}, \bibinfo{person}{Luke Metz},
  {and} \bibinfo{person}{Soumith Chintala}.} \bibinfo{year}{2015}\natexlab{}.
\newblock \showarticletitle{Unsupervised representation learning with deep
  convolutional generative adversarial networks}.
\newblock \bibinfo{journal}{\emph{arXiv preprint arXiv:1511.06434}}
  (\bibinfo{year}{2015}).
\newblock


\bibitem[\protect\citeauthoryear{Raval, Machanavajjhala, and Pan}{Raval
  et~al\mbox{.}}{2019}]%
        {bib:raval:pet2019}
\bibfield{author}{\bibinfo{person}{Nisarg Raval}, \bibinfo{person}{Ashwin
  Machanavajjhala}, {and} \bibinfo{person}{Jerry Pan}.}
  \bibinfo{year}{2019}\natexlab{}.
\newblock \showarticletitle{Olympus: sensor privacy through utility aware
  obfuscation}.
\newblock \bibinfo{journal}{\emph{Proceedings of PET}} (\bibinfo{year}{2019}).
\newblock


\bibitem[\protect\citeauthoryear{Sicong, Zimu, Junzhao, Longfei, Han, and
  Wang}{Sicong et~al\mbox{.}}{2017}]%
        {bib:sicong2017:IMWUT}
\bibfield{author}{\bibinfo{person}{Liu Sicong}, \bibinfo{person}{Zhou Zimu},
  \bibinfo{person}{Du Junzhao}, \bibinfo{person}{Shangguan Longfei},
  \bibinfo{person}{Jun Han}, {and} \bibinfo{person}{Xin Wang}.}
  \bibinfo{year}{2017}\natexlab{}.
\newblock \showarticletitle{UbiEar: Bringing Location-independent Sound
  Awareness to the Hard-of-hearing People with Smartphones}.
\newblock \bibinfo{journal}{\emph{Journal of IMWUT}} (\bibinfo{year}{2017}).
\newblock


\bibitem[\protect\citeauthoryear{team}{team}{2018}]%
        {lib:tf}
\bibfield{author}{\bibinfo{person}{Google~Brain team}.}
  \bibinfo{year}{2018}\natexlab{}.
\newblock \bibinfo{title}{TensorFlow}.
\newblock \bibinfo{howpublished}{\url{https://www.tensorflow.org/tutorials/}}.
\newblock


\bibitem[\protect\citeauthoryear{Truex, Baracaldo, Anwar, Steinke, Ludwig, and
  Zhang}{Truex et~al\mbox{.}}{2018}]%
        {bib:truex:arXiv2018}
\bibfield{author}{\bibinfo{person}{Stacey Truex}, \bibinfo{person}{Nathalie
  Baracaldo}, \bibinfo{person}{Ali Anwar}, \bibinfo{person}{Thomas Steinke},
  \bibinfo{person}{Heiko Ludwig}, {and} \bibinfo{person}{Rui Zhang}.}
  \bibinfo{year}{2018}\natexlab{}.
\newblock \showarticletitle{A hybrid approach to privacy-preserving federated
  learning}.
\newblock \bibinfo{journal}{\emph{arXiv preprint arXiv:1812.03224}}
  (\bibinfo{year}{2018}).
\newblock


\bibitem[\protect\citeauthoryear{UCI}{UCI}{2017}]%
        {data:Har}
\bibfield{author}{\bibinfo{person}{UCI}.} \bibinfo{year}{2017}\natexlab{}.
\newblock \bibinfo{title}{Har: Dataset for Human Activity Recognition}.
\newblock \bibinfo{howpublished}{\url{https://goo.gl/m5bRo1}}.
\newblock


\bibitem[\protect\citeauthoryear{Wu, Wang, Wang, and Jin}{Wu
  et~al\mbox{.}}{2018}]%
        {bib:wu:ECCV2018}
\bibfield{author}{\bibinfo{person}{Zhenyu Wu}, \bibinfo{person}{Zhangyang
  Wang}, \bibinfo{person}{Zhaowen Wang}, {and} \bibinfo{person}{Hailin Jin}.}
  \bibinfo{year}{2018}\natexlab{}.
\newblock \showarticletitle{Towards privacy-preserving visual recognition via
  adversarial training: A pilot study}. In
  \bibinfo{booktitle}{\emph{Proceedings of ECCV}}.
\newblock


\bibitem[\protect\citeauthoryear{Zeiler, Krishnan, Taylor, and Fergus}{Zeiler
  et~al\mbox{.}}{2010}]%
        {bib:zeiler2010:deconvolutional}
\bibfield{author}{\bibinfo{person}{Matthew~D Zeiler}, \bibinfo{person}{Dilip
  Krishnan}, \bibinfo{person}{Graham~W Taylor}, {and} \bibinfo{person}{Rob
  Fergus}.} \bibinfo{year}{2010}\natexlab{}.
\newblock \showarticletitle{Deconvolutional networks}. In
  \bibinfo{booktitle}{\emph{Proceedings of CVPR}}.
\newblock


\bibitem[\protect\citeauthoryear{Zhong, Wang, Ling, and Dong}{Zhong
  et~al\mbox{.}}{2016}]%
        {bib:zhong2016:FDS}
\bibfield{author}{\bibinfo{person}{Guoqiang Zhong}, \bibinfo{person}{Li-Na
  Wang}, \bibinfo{person}{Xiao Ling}, {and} \bibinfo{person}{Junyu Dong}.}
  \bibinfo{year}{2016}\natexlab{}.
\newblock \showarticletitle{An overview on data representation learning: From
  traditional feature learning to recent deep learning}.
\newblock \bibinfo{journal}{\emph{Journal of Finance and Data Science}}
  (\bibinfo{year}{2016}), \bibinfo{pages}{265--278}.
\newblock


\end{thebibliography}
